%% file: tikzero.tex
\pdfoutput=1
\PassOptionsToPackage{cachedir=minted,frozencache}{minted}
\documentclass[10pt,twocolumn,letterpaper]{article} 
\usepackage[pagenumbers]{styles/iccv}
\usepackage[pagebackref,breaklinks,colorlinks]{hyperref}
\usepackage{styles/tikzero}
\usepackage{booktabs,widetable,tabularx,subcaption}
\usepackage[most,minted]{tcolorbox}
\usetikzlibrary{
  calc,
  arrows.meta,
  decorations.pathmorphing,
  intersections,
  shapes,
  decorations.markings,
  patterns.meta
}

\graphicspath{{graphics}{graphics/examples}{graphics/openmoji}}
\makeatletter\let\input@path\Ginput@path\makeatother

\definecolor{iccvblue}{rgb}{0.21,0.49,0.74}
\hypersetup{%
  allcolors=iccvblue
}

\ifpreprint{}{\AtBeginDocument{\crefname{appendix}{Supp.\ Mat.}{Supp.\ Mats.}}}

\AddToHook{cmd/appendix/before}{\crefalias{section}{appendix}}

\makeatletter
\def\paragraph{%
  \@startsection{paragraph}%
    {4}{\z@}{6pt plus 2pt minus 2pt}{-1em}{\bf}%
}
\makeatother

\def\paperID{9782}
\def\confName{ICCV}
\def\confYear{2025}

\title{\projectname: Zero-Shot Text-Guided Graphics Program Synthesis}

\small
\newlength{\smallbaselineskip}
\normalsize

\author{
  Jonas Belouadi\footnotemark\qquad
  Eddy Ilg\footnotemark\qquad
  Margret Keuper\dblfootnotemark{1}{3}\qquad
  Hideki Tanaka\footnotemark[4]\qquad
  Masao Utiyama\footnotemark[4]\\
  Raj Dabre\footnotemark[4]\qquad
  Steffen Eger\footnotemark[2]\qquad
  Simone Ponzetto\footnotemark[1]\\
  \small University of Mannheim, Germany\footnotemark[1]\qquad
  \small University of Technology Nuremberg, Germany\footnotemark[2]\\[\dimexpr-\normalbaselineskip+\smallbaselineskip]
  \small Max Planck Institute for Informatics, Saarland Informatics Campus, Germany\footnotemark[3]\\[\dimexpr-\normalbaselineskip+\smallbaselineskip]
  \small National Institute of Information and Communications Technology, Japan\footnotemark[4]\\[\dimexpr-\normalbaselineskip+\smallbaselineskip]
  \small \mail{jonas.belouadi@uni-mannheim.de}%
}

\begin{document}

\maketitle

\begin{abstract}
  Automatically synthesizing figures from text captions is a compelling
  capability. However, achieving high geometric precision and editability
  requires representing figures as graphics programs in languages like
  \tikzname, and aligned training data (i.e., graphics programs with captions)
  remains scarce. Meanwhile, large amounts of unaligned graphics programs and
  captioned raster images are more readily available. We reconcile these
  disparate data sources by presenting \projectname, which decouples graphics
  program generation from text understanding by using image representations as
  an intermediary bridge. It enables independent training on graphics programs
  and captioned images and allows for zero-shot text-guided graphics program
  synthesis during inference. We show that our method substantially outperforms
  baselines that can only operate with caption-aligned graphics programs.
  Furthermore, when leveraging caption-aligned graphics programs as a
  complementary training signal, \projectname matches or exceeds the
  performance of much larger models, including commercial systems like \gpt.
  Our code, datasets, and select models \anonymize[will be made publicly
  available.]{are publicly
  available.\footnote{\url{https://github.com/potamides/DeTikZify}}}
\end{abstract}

\input{sections/introduction}
\input{sections/related}
\input{sections/method}
\input{sections/data_model}
\input{sections/experiments}
\input{sections/analysis}
\input{sections/conclusion}
\input{sections/limitations}
\anonymize[]{\input{sections/acknowledgements}}
{
    \small
    \bibliographystyle{styles/ieeenat_fullname}
    \bibliography{tikzero}
}

\clearpage\appendix
\input{sections/appendix}

\end{document}

%% file: sections/introduction.tex
\section{Introduction}\label{sec:introduction} 
\begin{figure}
  \centering
  \newcommand{\loadfig}[2][]{\raisebox{-.5\dimexpr\totalheight-\ht\strutbox}{\includegraphics[width=\linewidth,#1]{graphics/examples/qualitative/#2_pdf.pdf}}}
  \newcommand{\caploss}{\footnotesize 3D contour plot of a loss function.}
  \newcommand{\capmlp}{\footnotesize A multi-layer perceptron with two hidden layers.}
  \newcommand{\capgauss}{\footnotesize Gaussian probability density function (blue) with markers showing one standard deviation (red).}

  \begin{tabularx}{\columnwidth}{*{2}{X} | *{2}{X}} 
    \toprule
    \multicolumn{2}{c}{\thead{\automatikz[v2]\rlap{ \xmark}}} & \multicolumn{2}{c}{\thead{\projectname*\rlap{ \cmark}}}\\
    \cmidrule(lr){1-2}\cmidrule(lr){3-4}
    \loadfig{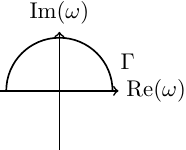} & \loadfig{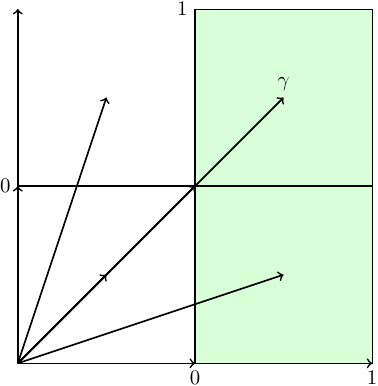} & \loadfig{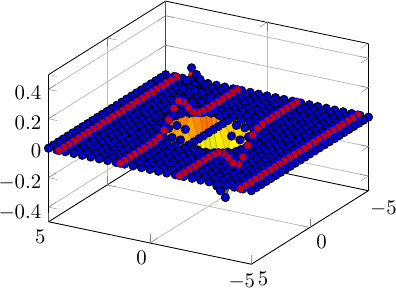} & \loadfig{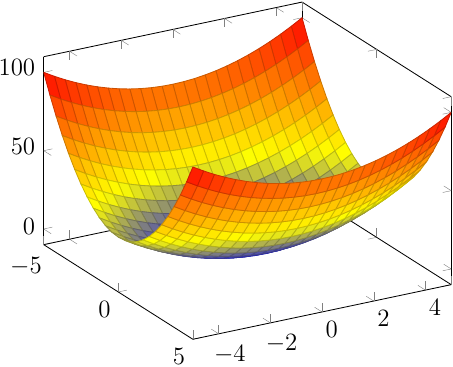}\\
    \multicolumn{4}{c}{\caploss}\\
    \cmidrule(lr){1-2}\cmidrule(lr){3-4}
    \loadfig{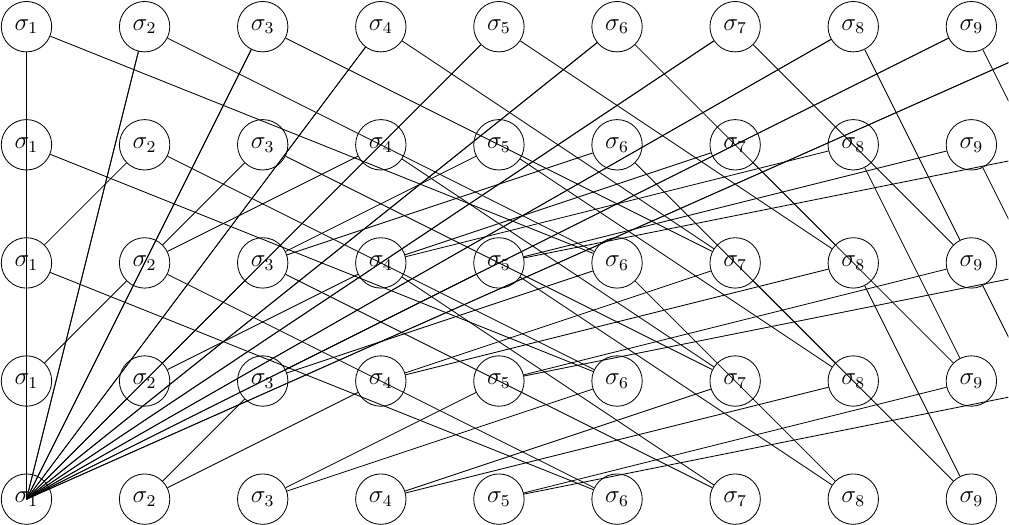} & \loadfig{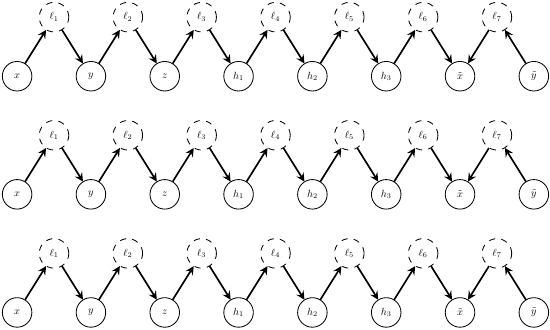} & \loadfig{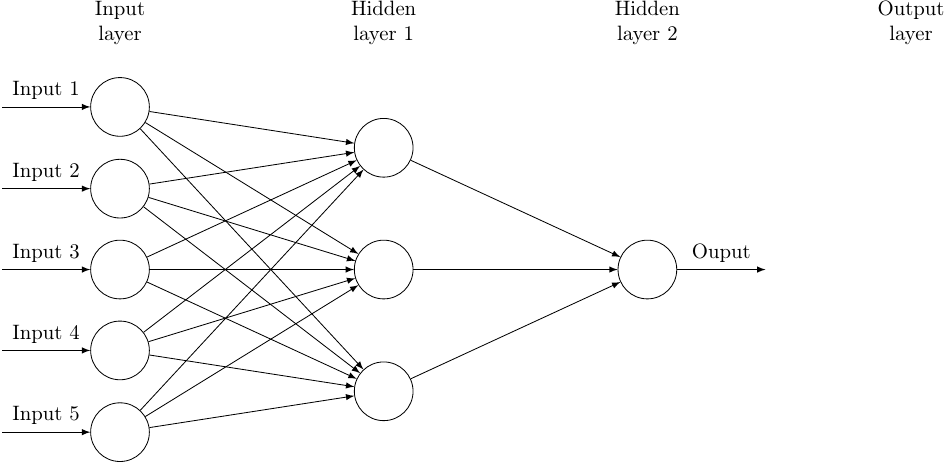} & \loadfig{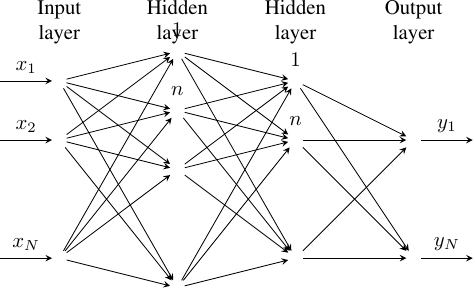}\\
    \multicolumn{4}{c}{\capmlp}\\
    \cmidrule(lr){1-2}\cmidrule(lr){3-4}
    \loadfig{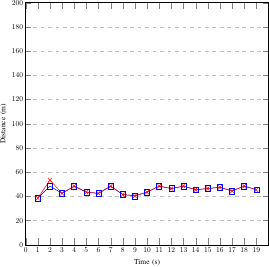} & \loadfig{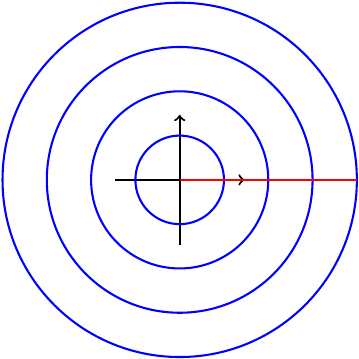} & \loadfig{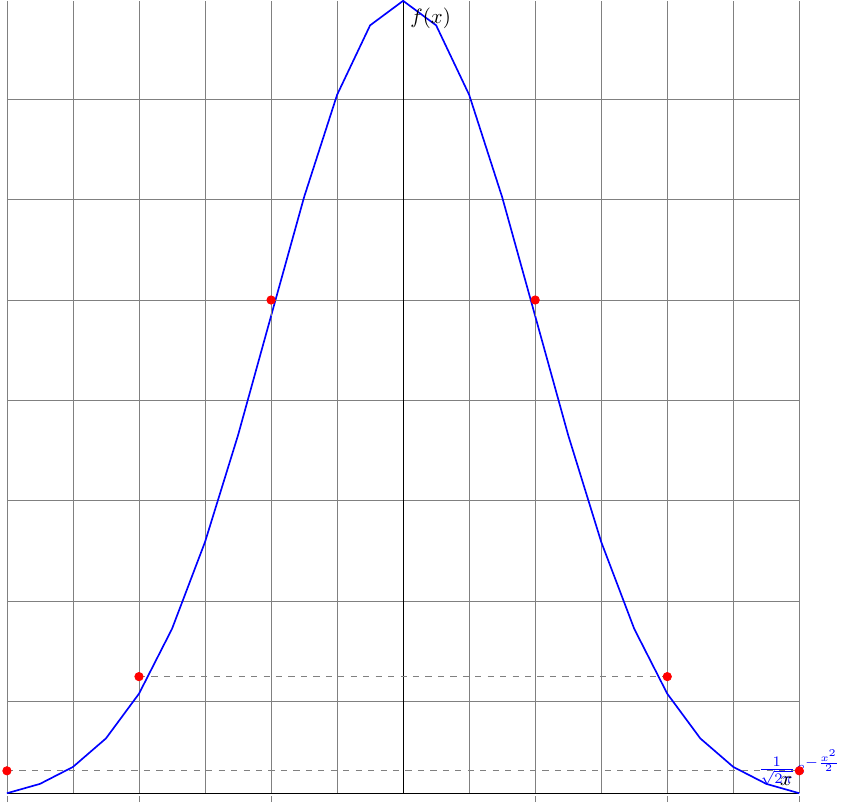} & \loadfig{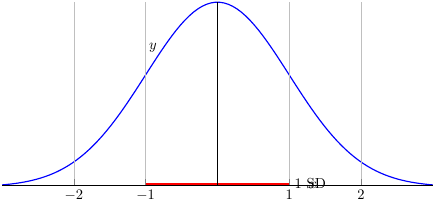}\\
    \multicolumn{4}{>{\hsize=\dimexpr4\hsize+6\tabcolsep+\arrayrulewidth\relax}X}{\centering\capgauss}\\
    \bottomrule
  \end{tabularx}
  \caption{Qualitative comparison of our \projectname* model (last two columns)
  and the end-to-end trained baseline \automatikz[v2][\llm; first two columns]
  on text-guided graphics program synthesis with \tikzname. Our method
  generates outputs that more closely follow the given captions. Example
  program listings are in \cref{sec:examples}.
  }%
  \label{fig:example}
\end{figure}
Graphics programming languages offer distinct advantages over low-level vector
formats (PDF, SVG) or raster image formats by representing visual concepts as
high-level programs that preserve semantics, remain human-interpretable, and
allow manual editing. These properties are particularly valuable in academia,
where specialized graphics programming languages like
\tikzname~\citep{tantau2023tikz} are popular for creating complex figures with
high expressivity. However, this comes with a steep learning curve, as seen on
the \tex Stack Exchange\footnote{\url{https://tex.stackexchange.com}} (\texse),
where nearly 10\% of questions concern \tikzname and make it the most
frequently discussed topic on the
platform~\citep{belouadi2024detikzify,belouadi2024automatikz}.

With recent advances in generative AI, simplifying the creation of graphics
programs has become increasingly feasible\@.  \citet{belouadi2024detikzify}
introduce \detikzify, an inverse graphics model that generates \tikzname
programs from images and hand-drawn sketches. However, creating these visual
inputs stays cumbersome, motivating alternative input modalities such as
natural language. While \citet{belouadi2024automatikz} propose \automatikz, a
text-guided synthesis model for \tikzname programs trained end-to-end on an
aligned caption-program corpus, its performance remains limited (cf.\
\cref{fig:example})~\citep{zhang2025scimage,zala2024diagrammergpt}.

\begin{figure}
  \centering
  \input{graphics/data.tex}%
  \caption{Illustration of training data availability for graphics program
  synthesis\@. \detikzify can leverage all graphics programs for training but
  lacks text guidance, while \automatikz is constrained to the small
  intersection of captioned graphics programs, resulting in limited
  performance. Our approach, \projectname, trains independently on both
  graphics programs and captioned images, enabling more effective use of
  available data and yielding superior results.}%
  \label{fig:training-data}
\end{figure}
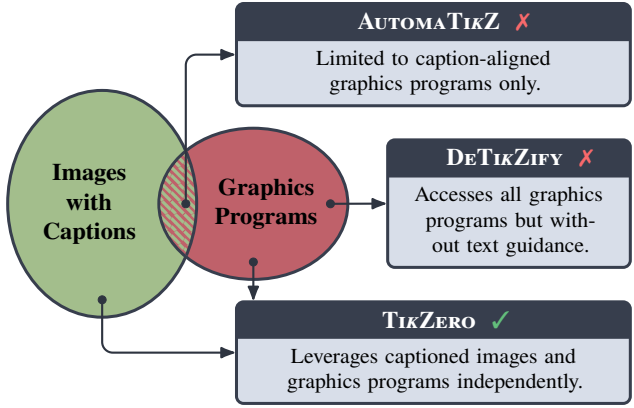
We identify insufficient training data as the primary limitation. Unlike
inverse graphics models such as \detikzify, which are inherently
self-supervised (trained by being conditioned on compiled representations of
their output programs) and can access sufficient training data (cf.\
\cref{fig:training-data}), end-to-end text-guided models like \automatikz
require graphics programs \emph{paired} with captions, substantially reducing
the available data pool (cf.\ \cref{fig:training-data}).

To address this challenge, we decouple the graphics program generation
component from text understanding, enabling independent training on graphics
programs and captioned images \emph{without} requiring paired data (cf.\
\cref{fig:training-data}). Our approach first trains an inverse graphics model
conditioned on image patch embeddings from a vision
encoder~\citep{radford2021learningtransferablevisualmodels}. We then train an
adapter network that generates synthetic image patch embeddings from captions.
This adapter training relies solely on captioned images, effectively
circumventing resource limitations and enabling zero-shot (in the sense that no
aligned caption-program examples are involved in the training process)
text-guided graphics program synthesis~\citep{palatucci2009zeroshot}. We
demonstrate that this approach, to which we refer as \projectname, outperforms
previous state-of-the-art methods (cf.\ \cref{fig:example}). Our key
contributions are:
\begin{enumerate}
  \item[(i)] A novel two-stage architecture, \projectname, which addresses the
    low-resource challenge in text-guided graphics program synthesis by
    aligning representation spaces rather than relying on aligned data.
  \item[(ii)] The \datikz[v3] dataset, comprising over 450k \tikzname graphics
    programs with roughly 170k captioned samples. Using this dataset, we
    train both \projectname and \automatikz[v2] (an updated version of
    \automatikz) on the same source data and show that \projectname
    outperforms \automatikz, \automatikz[v2], and other end-to-end trained
    baselines.
  \item[(iii)] An enhanced model, \projectname*, combining \projectname with
    the end-to-end fine-tuning of \automatikz[v2], which surpasses larger
    baselines and matches the performance of commercial models like
    \gpt~\citep{openai2023gpt4} on key metrics.
\end{enumerate}

%% file: graphics/data.tex
\newtcolorbox{system}[2][]{%
  enhanced,%
  halign title=center,%
  fonttitle=\small\bf,%
  fontupper=\footnotesize\normalfont,%
  halign=center,%
  colframe=nord3-dim,%
  colback=nord4,%
  colbacktitle=nord3-dim,%
  size=title,%
  left=0.25mm,%
  right=.25mm,%
  title={#2},%
  #1%
}

\tikzset{
    o/.style={
        decoration={
            markings,
            mark={
                at position 0
                with {
                  \filldraw[line width=0pt,nord0] circle [radius=#1/2];
                }
            }
        },
        postaction=decorate
    },
    titlewidth/.style={line width=0.4mm},
    o/.default=3pt
}

\begin{tikzpicture}[scale=0.5,thick,rounded corners,>={Latex[round]},font=\small\bf,align=center,draw=nord0,fill=nord0]
    \draw (-2,0) node[titlewidth,ellipse,draw=nord3-dim,fill=nord14,minimum width=2.5cm, minimum height=3cm, name path=captions] (captions) {};
    \draw (2,0) node[titlewidth,ellipse,draw=nord3-dim,fill=nord11,minimum width=2.5cm,minimum height=2cm,name path=programs] (programs) {};

    \begin{scope}
      \clip (programs) ellipse (2.5cm-.4mm and 2cm-.4mm);
      \draw[titlewidth,pattern={Lines[angle=-45, distance=.8mm,  line width=.4mm]}, pattern color=nord14,draw=nord3-dim] (captions) ellipse (2.5 and 3);
    \end{scope}

    \node[text width=1.5cm,xshift=.1cm] at ($(captions.west)!.5!(programs.west)$) {Images with Captions};
    \node[text width=1.5cm,xshift=-.1cm] at ($(captions.east)!.5!(programs.east)$) {Graphics Programs};

    \coordinate (top) at ($(programs.north)+(-.5,0.5)$);
    \coordinate (bot) at ($(programs.south)+(-.5,-0.5)$);
    \node[inner sep=0,anchor=south west] at (top) (automatikz) {%
      \begin{system}[width=5.25cm]{\automatikz\rlap{ \xmark}}%
        Limited to caption-aligned graphics programs only.
      \end{system}%
    };
    \node[inner sep=0,anchor=south west,anchor=north west] at (bot) (tikzero) {%
      \begin{system}[width=5.25cm]{\projectname\rlap{ \cmark}}%
        Leverages captioned images and graphics programs independently.
      \end{system}%
      };

    \node[inner sep=0,anchor=south west,anchor=east] at ($(automatikz.east)!.5!(tikzero.east)$) (detikzify) {%
      \begin{system}[width=3.25cm]{\detikzify\rlap{ \xmark}}%
        Accesses all graphics programs but without text guidance.
      \end{system}%
      };

    \draw[o,->,name intersections={of=captions and programs}] (intersection-1 |- programs) to [out=90,in=270] (intersection-1) |- (automatikz.west);
    \draw[o,->] ($(programs.east)-(.5cm,0)$) -- (detikzify.west);
    \draw[o,->] ($(captions.south)+(0,.5cm)$) |- (tikzero.west);
    \draw[o,->] ($(programs.south)+(0,.5cm)$) to (tikzero.north -| programs);
\end{tikzpicture}

%% file: sections/related.tex
\section{Related Work}\label{sec:related} 

\paragraph{Inverse Graphics Program Synthesis}
Inverse graphics, i.e., synthesizing a graphics program to reproduce a visual
target, represents a specialized instance of neural program
synthesis~\citep{parisotto2017neurosymbolic,devlin2017robustfill,ellis2021dreamcoder}.
Deep learning models have shown remarkable success in this
domain~\citep{ganin2018program,ellis2018drawing,ellis2019repl}, with \vlm*{}s
(\vlm{}s) increasingly gaining
prominence~\citep{belouadi2024detikzify,kulits2024rethinking,li2024is,kapur2025diffusion}.
While controlled experimental studies often rely on synthetic
datasets~\citep{kapur2025diffusion,kulits2024rethinking,ellis2018drawing,sharma2018csgnet,tian2018learning,amara2023uml},
real-world applications typically leverage more complex and diverse
human-created
data~\citep{belouadi2024detikzify,laurencon2024idefics2,laurencon2024idefics3,tong2024cambrian,rodriguez2023starvector,zhang2024mm15methodsanalysis},
highlighting the importance of data availability.
In scientific contexts, \tikzname has emerged as a popular choice due to its
versatility, expressiveness, and widespread adoption in academic
circles~\citep{belouadi2024detikzify,laurencon2024idefics2,laurencon2024idefics3,tong2024cambrian,zhang2024mm15methodsanalysis}.
Although these approaches are not tailored to text-guided generation, we
incorporate key elements from them into our approach.%

\paragraph{Text-Guided Graphics Program Synthesis}
Current text-guided approaches to graphics program synthesis remain limited,
mainly because of the scarcity of captioned graphics programs outlined in
\cref{sec:introduction}, but also because of the difficulty of generating
synthetic data with human-like
captions~\citep{li2024multimodalarxiv,kim2025multillmcollaborativecaptiongeneration}.
Researchers interested in this capability currently rely on the emerging
capabilities of large commercial models such as
\gpt~\citep{openai2023gpt4,bubeck2023sparks,sharma2024vision,zhang2023controllable},
which raises concerns about accessibility, reproducibility, and computational
cost~\citep{Chen2024How}. In contrast, related domains like vector graphics
generation~\citep{rodriguez2023starvector,polaczek2025neuralsvg,wu2023iconshop,Jain2023vectorfusion,frans2022clipdraw}
and
\nlvis~\citep{voigt2024plots,luo2021nl2vis,Mackinlay1986automating,roth1994interactive,wu2024nl2vis}
have shown more progress. Similar to inverse graphics, these fields
increasingly incorporate
\llm*{}s (\llm{}s)~\citep{wu2024nl2vis,voigt2024plots,rodriguez2023starvector}.
However, vector graphics approaches typically generate only low-level B\'ezier
curves, limiting output
complexity~\citep{polaczek2025neuralsvg,wu2023iconshop,belouadi2024detikzify},
and \nlvis focuses exclusively on data visualization with a restricted set of
visualization types~\citep{wu2024nl2vis}. More complex applications, such as
generating arbitrary scientific figures from captions with \tikzname, remain
underexplored---a gap we address in this work.%

\paragraph{Text-to-Image Generation}
\projectname shares conceptual and architectural similarities with several
text-to-image generation
methods~\citep{rombach2022highresolution,ramesh2021zeroshottexttoimagegeneration,ramesh2022dalle2,ding2021cogview,ding2022cogview2}.
\citet{rodriguez2023figgen,rodriguez2023ocrvqgan} explore generative
adversarial networks~\citep{goodfellow2020gan,esser2021vqgan} and diffusion
models~\citep{sohl2015diffusion,song2021denoising} for scientific figure
generation, but these approaches are tied to raster images, which are not ideal
for representing scientific figures\@. \citet{ramesh2022dalle2} propose a
two-stage model with independently trained prior and decoder components to
generate raster images from text. Although prior networks resemble our adapters
and have been used with inverse graphics models~\citep{hu2023procedural}, they
target \emph{global} image embeddings containing only abstract information,
which degrades performance when used with inverse graphics architectures that
work best with \emph{patch-level} details~\citep{yin2024mllm}. In contrast, our
adapters specifically operate on patch-level embeddings, and we demonstrate
that this \emph{improves} performance compared to end-to-end trained baselines.

%% file: sections/method.tex
\begin{figure*}
  \centering
  \input{graphics/architecture.tex}%
  \caption{
  Architecture overview of \projectname during inference. Solid lines represent
  the standard caption-conditioned path, which flows through the text encoder
  into the adapter network of \projectname before connecting to the vision
  encoder. In certain configurations (cf.\ \cref{sec:adapter-finetuning}), the
  caption also feeds into the text decoder (depicted by dotted lines and
  ``\textbullet'' markers representing shortcuts). The self- and
  cross-attention layers (\textcolor{nord13-dim}{yellow}) are simplified
  representations, omitting internal feed-forward layers and residual
  connections~\citep{he2016resnet}. An exception is the explicit residual
  connection between the cross-attention and self-attention layers of the
  vision encoder, visualizing the gating mechanism $\gamma$
  (\textcolor{nord15-dim}{purple}). Additionally, the dashed path illustrates
  how the inverse graphics model generates graphics programs when conditioned
  on images.
  }%
  \label{fig:architecture}
\end{figure*}
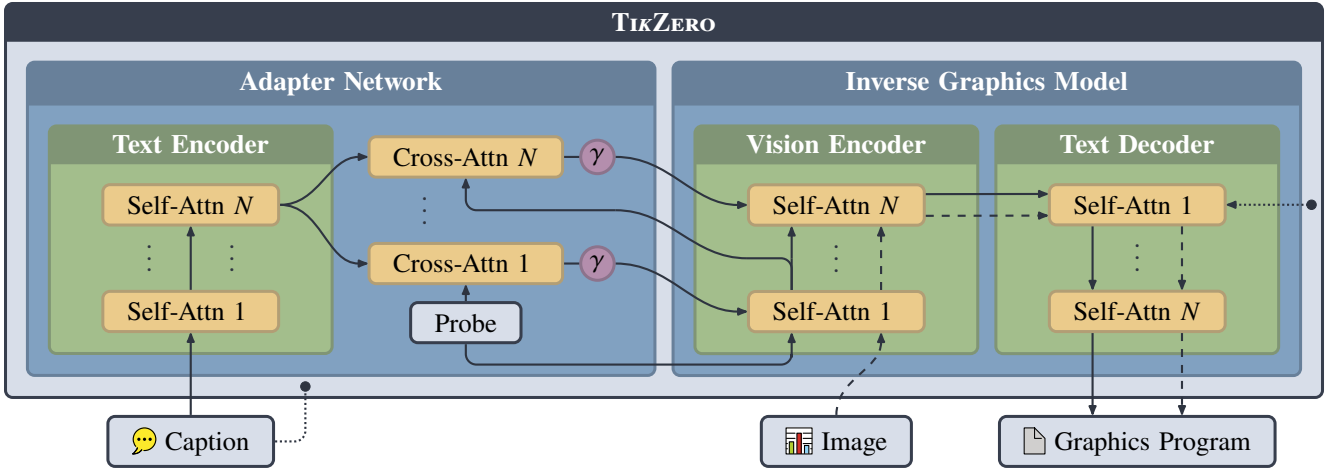
\section{The \projectname{} Model \& Architecture}\label{sec:method}
As the foundation of \projectname, we first develop a state-of-the-art inverse
graphics model for graphics program synthesis. We then incorporate a
cross-attention adapter network~\citep{vaswani2017attention} for text
guidance\@. \cref{fig:architecture} provides an overview of our method.

\paragraph{The Inverse Graphics Model}
Due to their demonstrated effectiveness (cf.\ \cref{sec:related}), we adopt a
\vlm architecture for the inverse graphics model of \projectname. 
\Cref{fig:architecture} illustrates its inner workings (dashed lines): the
model processes rasterized images and autoregressively generates their
corresponding programs without involving captions at this stage.

\paragraph{The Adapter Network}
\vlm{}s consist of two primary components: a vision encoder that produces image
embeddings and a text decoder that, in our case, generates graphics programs
conditioned on these embeddings. The unidirectional and localized flow of
information between these components allows us to inject additional information
solely into the vision encoder, thereby influencing the output of the text
decoder. We exploit this property by introducing a trainable, text-conditioned
adapter network that mimics the outputs of the original vision encoder.
This effectively enables zero-shot generation of graphics programs conditioned
on text when its outputs are fed into the decoder. In addition to
circumventing the resource limitations discussed in \cref{sec:introduction},
this architecture has other welcome implications: During adapter training, the
text decoder, usually the largest component of the model, does not need to be
loaded, resulting in efficient and fast training even with large datasets.
Our adapter incorporates a lightweight text encoder for embedding captions and
introduces newly initialized gated cross-attention
layers~\citep{alayrac2022flamingo,grattafiori2024llama3herdmodels} before each
vision encoder layer (cf., \cref{fig:architecture}). The keys and values derive
from the final text encoder representations, while the queries originate from a
trainable probe used instead of image inputs. The gates $\gamma$ allow the
model to learn at which layers and to what extent information from the text
encoder should flow into the vision encoder. Contrary to existing literature,
which often employs $\tanh$ gates~\citep{hochreiter1997lstm} that initialize to
zero (indicating no information flow), we find that using
$\operatorname{sigmoid}$ gates (0.5 at initialization) accelerates training
convergence since, in our case, only little information originates from the
vision encoder inputs (i.e., the probe). We ablate the gates and the probe in
\cref{sec:ablations}.

\paragraph{Training Objective}
Given a caption-image dataset for training, we first embed the patches $p \in
\dlvec{p}$ of image $i$ using the unmodified vision encoder $\operatorname{M}$
of our \vlm. Subsequently, we incorporate the cross-attention adapter to obtain
the modified encoder $\operatorname{\widehat{M}}$, which we then distill on
these image patch embeddings conditioned solely on the caption $t$ and probe
$\hat{\imath}$~\citep{hinton2015distillingknowledgeneuralnetwork}. This leads
to the following objective:
\begin{equation}
  \mathcal{L}_{\text{dist}} =%
    \frac{1}{\norm{\dlvec{p}}}\sum_{p \in \dlvec{p}}%
    \dist*{%
      \model[_\theta]*{\,p\mid i\,},%
      \hatmodel[_{\theta,\hat{\theta}}]*{\,p\mid \hat{\imath},t\,}%
    },
\end{equation}
where $\dist{\dlvec{x},\dlvec{y}}$ represents a distance metric. Following
common practices in model
distillation~\citep{Yang2024clipkd,sanh2020distilbert}, we experiment with
\cosdist* and \mse*. Here, $\theta$ denotes the original model parameters that
remain fully frozen, while $\hat{\theta}$ represents the adapter parameters of
which the cross-attention layers and the image probe are trainable.

%% file: graphics/architecture.tex
\newtcbox{\layer}[1][]{%
  enhanced,%
  size=title,%
  colframe=nord13-dim,%
  colback=nord13,%
  #1%
}
\newtcbox{\probe}[1][]{%
  enhanced,%
  size=title,%
  colframe=nord3-dim,%
  colback=nord4,%
  #1%
}
\newcommand{\gate}[1][]{%
  \layer[colframe=nord15-dim,colback=nord15,left=0pt,right=0pt,circular arc,square,valign=center,#1]{$\gamma$}%
}
\newcommand{\selfattn}[2][]{%
  \layer[#1]{Self-Attn \rlap{$#2$}\hphantom{$N$}}%
}
\newcommand{\crossattn}[2][]{%
  \layer[#1]{Cross-Attn \rlap{$#2$}\hphantom{$N$}}%
}
\newcommand{\addicon}[1]{%
  \raisebox{-.15\baselineskip}{\includegraphics[height=.85\baselineskip]{graphics/openmoji/#1.pdf}}%
}
\newcommand{\inputs}[3][]{%
  \probe[#1]{\addicon{#2} #3}
}
\newtcolorbox{system}[2][]{%
  enhanced,%
  halign title=center,%
  halign=center,%
  size=title,%
  left=1.5mm,%
  right=1.5mm,%
  top=1.5mm,%
  bottom=1.5mm,%
  title={#2},%
  #1%
}
\newcommand{\cvdots}{%
  \tcbox[size=minimal,frame empty,opacityback=0,fontupper=\color{nord0},valign=center]{%
    \rotatebox{90}{$\ldots$}%
  }%
}
\newcommand{\lrvdots}{%
  \tcbox[size=minimal,frame empty,opacityback=0,fontupper=\color{nord0},valign=center]{%
    \rotatebox{90}{$\ldots$}\quad\qquad\rotatebox{90}{$\ldots$}%
  }%
}
\newcommand{\lvdots}{%
  \tcbox[size=minimal,frame empty,opacityback=0,fontupper=\color{nord0},valign=center]{%
    \rotatebox{90}{$\ldots$}\quad\qquad\rotatebox{90}{\vphantom{$\ldots$}}%
  }%
}

\begin{system}[remember as=detikzify-adp,colframe=nord3-dim,colback=nord4]{\bf\projectname}
  \begin{tcbraster}[raster columns=4,raster valign=top]
    \begin{system}[remember as=adapter,raster multicolumn=2,colframe=nord9-dim,colback=nord9]{\bf Adapter Network}
      \begin{tcbraster}[raster columns=2, raster equal height]
        \begin{system}[remember as=text-encoder,colframe=nord14-dim,colback=nord14]{\bf Text Encoder}
          \selfattn[remember as=text-attn-n]{N}
          \lrvdots
          \selfattn[remember as=text-attn-1]{1}
        \end{system}
        \begin{system}[valign=center,colframe=nord11,colback=nord11,bottom=0pt,left=0pt,right=0pt,blankest]{\bf\vphantom{Text Encoder}}
        \begin{tcbraster}[raster before skip=0pt,raster row skip=6pt,raster columns=2,raster force size=false,raster halign=center,raster valign=center,raster equal height=none]
          \crossattn[remember as=cross-attn-n]{N}\gate[remember as=gate-n]
          \lvdots\gate[height=0pt,frame empty,invisible,opacityback=0]
          \crossattn[remember as=cross-attn-1]{1}\gate[remember as=gate-1]
          \probe[remember as=probe]{Probe}\gate[frame empty,invisible,opacityback=0]
        \end{tcbraster}
        \end{system}
      \end{tcbraster}
    \end{system}
    \begin{system}[remember as=detikzify,raster multicolumn=2,colframe=nord9-dim,colback=nord9]{\bf Inverse Graphics Model}
      \begin{tcbraster}[raster columns=2, raster equal height]
        \begin{system}[remember as=vision-encoder,colframe=nord14-dim,colback=nord14]{\bf Vision Encoder}
          \selfattn[remember as=vision-attn-n]{N}
          \cvdots
          \selfattn[remember as=vision-attn-1]{1}
        \end{system}
        \begin{system}[remember as=text-decoder, colframe=nord14-dim,colback=nord14]{\bf Text Decoder}
          \selfattn[remember as=code-attn-1]{1}
          \cvdots
          \selfattn[remember as=code-attn-n]{N}
        \end{system}
      \end{tcbraster}
    \end{system}
  \end{tcbraster}
\end{system}
\begin{tcbraster}[raster left skip=5.8mm,raster right skip=5.8mm,raster column skip=7.8mm,raster columns=4, raster equal height]
  \begin{tcboxedraster}[raster left skip=0pt, raster right skip=0pt,raster column skip=2mm,raster columns=2, raster equal height,size=title]{raster multicolumn=2,blankest}
    \tcbox[blankest,halign=center]{\inputs[remember as=caption]{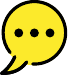}{Caption}}
    \tcbox[blankest]{}
  \end{tcboxedraster}
  \begin{tcboxedraster}[raster left skip=0pt, raster right skip=0pt,raster column skip=2mm,raster columns=2, raster equal height,size=title]{raster multicolumn=2,blankest}
    \tcbox[blankest,halign=center]{\inputs[remember as=image]{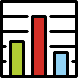}{Image}}
    \tcbox[blankest,halign=center]{\inputs[remember as=program]{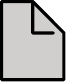}{Graphics Program}}
  \end{tcboxedraster}
  \begin{tikzpicture}[overlay,remember picture,thick,rounded corners,>={Latex[round,length=1.75mm]},draw=nord0,fill=nord0]%
    \draw[->] (caption.north -| text-attn-1.south) to (text-attn-1.south);
    \draw[->] (text-attn-1.north) to (text-attn-n.south);

    \draw[->] (text-attn-n.east) to[out=0,in=180] (cross-attn-1.west);
    \draw[->] (text-attn-n.east) to[out=0,in=180] (cross-attn-n.west);
    \draw[->] (probe.north) to (cross-attn-1.south);
    \draw[-] (cross-attn-1.east) to (gate-1.west);
    \draw[-] (cross-attn-n.east) to (gate-n.west);

    \draw[->] (gate-n.east) to[out=0,in=180] (vision-attn-n.west);
    \draw[->] (gate-1.east) to[out=0,in=180] (vision-attn-1.west);
    \draw[->,dashed] (image.north) to[out=90,in=270] ($(vision-attn-1.south)!.5!(vision-attn-1.south east)$);

    \coordinate (probe-input) at ($(vision-attn-1.south)!.5!(vision-attn-1.south west)$);
    \coordinate (probe-tmp) at ($(vision-encoder.south)!.5!(detikzify.south)+(0,.2mm)$);
    \draw[->] (probe.south) |- (probe-tmp -| probe-input) -| (probe-input);

    \draw[-] ($(vision-attn-1.north)!.5!(vision-attn-1.north west)$) to ({$(vision-attn-1.north)!.5!(vision-attn-n.south)$} -| {$(vision-attn-n.south)!.5!(vision-attn-n.south west)$})
      {[->]
      edge ($(vision-attn-n.south)!.5!(vision-attn-n.south west)$)
      };

    \draw[-] ($(vision-attn-1.north)!.5!(vision-attn-1.north west)$) |- ($(vision-attn-1.west)!.5!(vision-attn-n.west)$);
    \draw[-] ($(vision-attn-1.west)!.5!(vision-attn-n.west)$) to[out=180,in=0] ($(gate-1.east)!.5!(gate-n.east)$);
    \draw[->] ($(gate-1.east)!.5!(gate-n.east)$) -| (cross-attn-n.south);
    \draw[->,dashed] ($(vision-attn-1.north)!.5!(vision-attn-1.north east)$) to ($(vision-attn-n.south)!.5!(vision-attn-n.south east)$);

    \draw[->] ($(vision-attn-n.east)!.5!(vision-attn-n.north east)$) to ($(code-attn-1.west)!.5!(code-attn-1.north west)$);
    \draw[->] ($(code-attn-1.south)!.5!(code-attn-1.south west)$) to ($(code-attn-n.north)!.5!(code-attn-n.north west)$);
    \draw[->] ($(code-attn-n.south)!.5!(code-attn-n.south west)$) to ({$(code-attn-n.south)!.5!(code-attn-n.south west)$} |- {$(program.north)$});
    \draw[->,dashed] ($(code-attn-1.south)!.5!(code-attn-1.south east)$) to ($(code-attn-n.north)!.5!(code-attn-n.north east)$);
    \draw[->,dashed] ($(vision-attn-n.east)!.5!(vision-attn-n.south east)$) to ($(code-attn-1.west)!.5!(code-attn-1.south west)$);
    \draw[->,dashed] ($(code-attn-n.south)!.5!(code-attn-n.south east)$) to ({$(code-attn-n.south)!.5!(code-attn-n.south east)$} |- {$(program.north)$});

    \coordinate (portal-start) at ($(detikzify.south)!.5!(detikzify-adp.south)+(0,.2mm)$);
    \coordinate (portal-end) at ($(text-encoder.east)!.5!(text-attn-n.east)-(.2mm,0)$);
    \draw[densely dotted] (caption.east) -| (portal-end |- portal-start);
    \filldraw[line width=0pt] (portal-end |- portal-start) circle [radius=1.25mm/2];

    \draw[->,densely dotted] ({$(detikzify.east)!.5!(detikzify-adp.east)$} |- {$(code-attn-1.east)$}) to (code-attn-1.east);
    \filldraw[line width=0pt] ({$(detikzify.east)!.5!(detikzify-adp.east)$} |- {$(code-attn-1.east)$}) circle [xshift=-.2mm,radius=1.25mm/2];
  \end{tikzpicture}
\end{tcbraster}

%% file: sections/data_model.tex
\section{Datasets \& Model Training}\label{sec:data_model} 
We introduce \datikz[v3], a novel dataset of \tikzname graphics programs
designed to support the training and evaluation of \projectname. Additionally,
we train \automatikz[v2] as a directly comparable baseline operating on the
same data source.

\begin{table}
  \centering
  \begin{tabular}{l *{3}{s{3.3}}}
    \toprule
    \thead{Source} & \nhead{\datikz} & \nhead{\datikz[v2]} & \nhead{\datikz[v3]}\\
    \midrule
    curated    &    ,981 &   1,566 &   3,646\\
    \texse     &  29,238 &  30,609 &  42,654\\
    \arxiv     &  85,656 & 326,450 & 407,851\\
    artificial &   1,957 &   1,958 &   2,256\\
    \midrule
    all        & 117,832 & 360,583 & 456,469\\
    \bottomrule
  \end{tabular}
  \caption{Breakdown of the number of unique \tikzname graphics in \datikz[v3]
  compared to its predecessors \datikz and \datikz[v2]. Qualitative examples
  can be found in \cref{sec:examples}.}%
  \label{tab:datikz-dataset}
\end{table}
\paragraph{The \datikz[v3] Dataset}
\datikz[v3] expands upon its predecessors \datikz and
\datikz[v2]~\citep{belouadi2024automatikz,belouadi2024detikzify}, incorporating
programs from curated repositories, \texse, \arxiv papers, and artificial
samples (cf.\ \cref{tab:datikz-dataset}). While previous versions focused
exclusively on \tikzname graphics with (v1) or without (v2) captions,
\datikz[v3] systematically extracts captions alongside \tikzname graphics
whenever possible to support our claims. From over 450k instances, fewer than
170k include captions, underscoring the challenges discussed in
\cref{sec:introduction}.

\paragraph{Training \projectname}
\projectname's \vlm builds upon \detikzify~\citep{belouadi2024detikzify} by
conditioning a \llama-based text decoder~\citep{touvron2023llama} on patch
embeddings from a \siglip vision encoder~\citep{zhai2023siglip}. Specifically,
we combine \llama[3.1] (8B)~\citep{grattafiori2024llama3herdmodels} with
\siglip \sovit. Unlike \detikzify and inspired by the continued \vit
pretraining approach of \internvl~\citep{chen2024fargpt4vclosinggap}, we
initialize the vision encoder with weights from the fine-tuned encoder of
\paligemma~\citep{beyer2024paligemmaversatile3bvlm} and increase the input
resolution to $420 \times 420$ pixels. Furthermore, we fully fine-tune the
vision encoder alongside the rest of the model instead of freezing it. We train
on \datikz[v3] for 5 epochs with a learning rate of \lr{5}{5} and a batch size
of 128\@. \projectname's \vlm consistently outperforms \detikzify, with
detailed evaluation results provided in \cref{sec:detikzify-details}.
For the adapter network, we initialize with \llama[3.2] (1B) as the text
encoder~\citep{grattafiori2024llama3herdmodels} and leverage
\arxivcap~\citep{li2024multimodalarxiv}, a dataset comprising 6.4 million
scientific caption-image pairs for training. The adapter accounts for 2 billion
of \projectname's 10 billion total parameters, with only 400 million being
trainable. We train for 3 epochs with a learning rate of \lr{1}{4} and a batch
size of 512. We emphasize that this two-stage training process \emph{does not}
access caption-program pairs. However, we demonstrate that incorporating such
aligned data in a subsequent fine-tuning step (\cref{sec:adapter-finetuning})
further enhances performance.

\paragraph{Training \automatikz[v2]}
Similar to its predecessor, \automatikz[v2] is a \emph{token-conditioned} \llm
that uses \emph{tokenized} captions as conditioning information for graphics
prediction (rather than patch embeddings). We initialize \automatikz[v2] in two
different ways: (i) \automatikz[v2][\llm], which starts from vanilla
\llama[3.1][8B] weights, and (ii) \automatikz[v2][\vlm], which leverages
\projectname's trained \vlm (minus the vision encoder) to benefit from transfer
learning~\citep{zhuang2020comprehensivesurveytransferlearning} on its larger
training corpus. Both variants employ the same hyperparameters as
\projectname's \vlm but can only utilize the caption-annotated subset of
\datikz[v3] for training. Despite having access to less caption-aligned data
than \projectname's adapter network, \automatikz[v2] requires a longer training
period primarily due to fine-tuning the large decoder (8 billion trainable
parameters versus the adapter network's 400 million). Training requires more
than two days for \automatikz[v2] and 1.5 days for \projectname's adapter
network when using eight Nvidia A100 40GB GPUs.

%% file: sections/experiments.tex
\section{Experiments}\label{sec:experiments} 
Before training models on \datikz[v3], we extract 1k samples from its captioned
subset to form our test set. To mitigate data leakage from pretraining to
testing, we only include instances created after the cut-off date specified by
\llama[3.2] and \arxivcap. We also employ an \ngram matching algorithm to avoid
cross-contamination with our training split~\citep{openai2023gpt4}. For all
models, the temperature is set to 0.8 and top-p to 0.95. Example outputs are
provided in \cref{fig:example} and \cref{sec:examples}.

\begin{table*}
  \centering

  \newcolumntype{\dsimcol}[1]{>{\collectcell{\gradcol{#1}{52.829}{52.024}{38.313}}}c<{\endcollectcell}}
  \newcolumntype{\kidcol}[1]{>{\collectcell{\gradcol{#1}{1.294}{3.491}{33.203}}}c<{\endcollectcell}}
  \newcolumntype{\clipcol}[1]{>{\collectcell{\gradcol{#1}{15.766}{14.327}{0.775}}}c<{\endcollectcell}}
  \newcolumntype{\cbleucol}[1]{>{\collectcell{\gradcol{#1}{1.723}{1.603}{0.328}}}c<{\endcollectcell}}
  \newcolumntype{\tedcol}[1]{>{\collectcell{\gradcol{#1}{62.24}{62.307}{76.985}}}c<{\endcollectcell}}
  \newcolumntype{\mtecol}[1]{>{\collectcell{\gradcol{#1}{85.866}{82.291}{21.595}}}c<{\endcollectcell}}
  \newcolumntype{\avgcol}[1]{>{\collectcell{\gradcol{#1}{85.599}{85.004}{0.0}}}c<{\endcollectcell}}
  \newcolumntype{\rclipcol}[1]{>{\collectcell{\gradcol{#1}{8.237}{8.002}{0.284}}}c<{\endcollectcell}}
  \newcolumntype{\ratiocol}[1]{>{\collectcell{\gradcol{#1}{77.831}{74.975}{33.858}}}c<{\endcollectcell}}

  \setlength{\extrarowheight}{\belowrulesep}\setlength{\belowrulesep}{0pt}
  \begin{tabular}{l
    \dsimcol{2.3} \kidcol{2.3} \clipcol{2.3} \cbleucol{1.3} \tedcol{2.3} \mtecol{2.3} \avgcol{2.3} \rclipcol{1.3} \ratiocol{2.3}}
    \toprule
    & \multicolumn{7}{c}{\thead{Original Text}} & \multicolumn{2}{c}{\thead{Redacted Text}}\\
    \cmidrule(l{\tabcolsep}r{\tabcolsep}){2-8}\cmidrule(l{\tabcolsep}r{\tabcolsep}){9-10}
    \thead{Models} & \nhead{\dsim\up} & \nhead{\kid\down} & \nhead{\clip\up} & \nhead{\cbleu\up} & \nhead{\ted\down} & \nhead{\mte\up} & \nhead{\underline{\avg}\up} & \nhead{\clip\up} & \nhead{Ratio\up}\\
    \midrule
    \idefics           & 45.475 & 11.426 & 14.327 & 0.656 & 63.175 & 69.558 & 66.628 & 4.851 & 33.858\\
    \automatikz[][13B] & 46.033 &  1.294 &  3.955 & 0.386 & 62.24  & 85.866 & 63.093 & 2.965 & 74.975\\[-\aboverulesep]
    \midrule
    \automatikz[v2][\vlm] & 38.313 & 33.203 &  0.775 &  0.328 & 76.985 & 21.595 & 0.0    & 0.284 & 36.597\\
    \automatikz[v2][\llm] & 50.548 &  3.491 & 15.766 &  0.658 & 62.307 & 81.775 & 82.375 & 8.002 & 50.753\\
    \projectname[MSE]     & 52.024 &  5.664 & 10.583 &  1.723 & 66.07  & 79.318 & 85.004 & 8.237 & 77.831\\
    \projectname[Cos]     & 52.829 &  5.103 & 10.051 &  1.603 & 65.51  & 82.291 & 85.599 & 7.226 & 71.893\\[-\aboverulesep]
    \bottomrule
  \end{tabular}
  \caption{System-level $\text{scores}\times100$ for \projectname and baselines
  of comparable size and training setup. Bold and underlined values denote the
  best and second-best scores for each metric column, respectively. Cell
  shading illustrates relative score magnitudes. Arrows indicate metric
  directionality. Overall, \projectname achieves the strongest average
  performance across metrics.}%
  \label{tab:adapter-vs-e2e}
\end{table*}
\paragraph{Evaluation Metrics} The multimodal nature of our task allows for
various evaluation metrics in our automatic evaluations. We assess perceptual
\emph{image similarity} between generated outputs and references by computing
\dsim* (\thead{\dsim})~\citep{fu2023dreamsim,sundaram2024when}, which
correlates highly with human judgments for scientific
images~\citep{belouadi2024detikzify}. We also calculate the \kid*
(\thead{\kid})~\citep{bińkowski2018kid} using \siglip image features, which
evaluates the overall quality of generated figures by comparing to the
distribution of reference figures.
We evaluate \emph{caption similarity} between generated outputs and reference
captions using \clip* (\thead{\clip})~\citep{hessel2021clipscore} with \siglip
features.
To measure \emph{code similarity} between generated and reference \tikzname
programs, we use \cbleu* (\thead{\cbleu}), a \bleu variant optimized for code
evaluation~\citep{eghbali2023bleu,papineni2002bleu}, and \ted*
(\thead{\ted})~\citep{belouadi2024detikzify}, a variant of the
\eed*~\citep{stanchev2019eed} utilizing a \tex tokenizer.
Since some metrics require that generated programs compile to images,
resampling is necessary if the output contains irrecoverable errors. To
quantify this, we compute the \emph{\mte*} (\thead{\mte}), defined as the 10\%
winsorized mean of the ratio between the number of tokens in the final
\tikzname program and the total number of tokens generated to produce that
program.
For a comprehensive view of model performance, we calculate the arithmetic mean
(\thead{AVG}) of \emph{all} previous metrics. As these metrics operate on
different scales, we apply min-max normalization before computing the average.
Additionally, some metrics are recomputed with redacted text in the outputs as
part of our analysis, cf.\ \cref{sec:clip-matching}.

\begin{table*}
  \centering

  \newcolumntype{\dsimcol}[1]{>{\collectcell{\gradcol{#1}{56.464}{56.295}{52.829}}}c<{\endcollectcell}}
  \newcolumntype{\kidcol}[1]{>{\collectcell{\gradcol{#1}{1.794}{1.831}{5.493}}}c<{\endcollectcell}}
  \newcolumntype{\clipcol}[1]{>{\collectcell{\gradcol{#1}{31.787}{24.87}{10.051}}}c<{\endcollectcell}}
  \newcolumntype{\cbleucol}[1]{>{\collectcell{\gradcol{#1}{1.988}{1.603}{0.285}}}c<{\endcollectcell}}
  \newcolumntype{\tedcol}[1]{>{\collectcell{\gradcol{#1}{58.511}{59.008}{65.51}}}c<{\endcollectcell}}
  \newcolumntype{\mtecol}[1]{>{\collectcell{\gradcol{#1}{97.675}{97.269}{82.291}}}c<{\endcollectcell}}
  \newcolumntype{\avgcol}[1]{>{\collectcell{\gradcol{#1}{87.043}{79.019}{14.658}}}c<{\endcollectcell}}
  \newcolumntype{\rclipcol}[1]{>{\collectcell{\gradcol{#1}{13.32}{12.164}{6.512}}}c<{\endcollectcell}}
  \newcolumntype{\ratiocol}[1]{>{\collectcell{\gradcol{#1}{71.893}{60.931}{41.905}}}c<{\endcollectcell}}

  \newcommand{\plus}[1]{$\phantom{}+\textrm{#1}$}
  \setlength{\extrarowheight}{\belowrulesep}\setlength{\belowrulesep}{0pt}
  \begin{tabular}{l
    \dsimcol{2.3} \kidcol{1.3} \clipcol{2.3} \cbleucol{1.3} \tedcol{2.3} \mtecol{2.3} \avgcol{2.3} \rclipcol{1.3} \ratiocol{2.3}}
    \toprule
    & \multicolumn{7}{c}{\thead{Original Text}} & \multicolumn{2}{c}{\thead{Redacted Text}}\\
    \cmidrule(l{\tabcolsep}r{\tabcolsep}){2-8}\cmidrule(l{\tabcolsep}r{\tabcolsep}){9-10}
    \thead{Models} & \nhead{\dsim\up} & \nhead{\kid\down} & \nhead{\clip\up} & \nhead{\cbleu\up} & \nhead{\ted\down} & \nhead{\mte\up} & \nhead{\underline{\avg}\up} & \nhead{\clip\up} & \nhead{Ratio\up}\\
    \midrule
    \qwen & 54.473 & 5.493 & 24.87  & 0.285 & 59.856 & 97.269 & 48.593 & 12.164 & 48.911\\
    \gpt  & 56.464 & 2.844 & 31.787 & 0.327 & 58.511 & 97.675 & 79.019 & 13.32  & 41.905\\[-\aboverulesep]
    \midrule
    \projectname[Cos]          & 52.829 & 5.103 & 10.051 & 1.603 & 65.51  & 82.291 & 14.658 &  7.226 & 71.893\\
    \plus{Fine-tuning}~(i)        & 53.203 & 1.794 & 10.687 & 0.759 & 61.572 & 94.851 & 46.497 &  6.512 & 60.931\\
    \plus{Separate Captions}~(ii) & 52.983 & 2.905 & 15.72  & 0.804 & 61.32  & 95.722 & 46.326 &  8.741 & 55.608\\
    \plus{Weight Resetting}~(iii) & 56.295 & 1.831 & 24.177 & 1.988 & 59.008 & 93.058 & 87.043 & 11.479 & 47.478\\[-\aboverulesep]
    \bottomrule
  \end{tabular}
  \caption{System-level $\text{scores}\times 100$ for additional baselines and
  \projectname combined with fine-tuning and token-conditioning. The scores for
  \projectname[Cos] are replicated from \cref{tab:adapter-vs-e2e} for
  convenience. Bold and underlined values denote the best and second-best
  scores for each metric column, respectively. Cell shading illustrates
  relative score magnitudes. Arrows indicate metric directionality. Overall,
  \projectname[Cos] with Weight Resetting (iii) demonstrates the strongest
  average performance across metrics.}%
  \label{tab:adapter-ft}
\end{table*}
\subsection{Comparison against End-to-End Fine-Tuning}\label{sec:adapter-vs-e2e}
In our initial experiment, we evaluate the zero-shot performance of
\projectname, trained as described in \cref{sec:method,sec:data_model} using
either \cosdist* (\cosdist) or \mse* (\mse), and compare it against end-to-end
trained baselines.

\paragraph{Baselines} Besides \automatikz[v2][\llm \& \vlm], which we designed
as directly comparable baselines, we assess other token-conditioned models of
similar and slightly larger sizes trained on \tikzname. Specifically,
we evaluate
\automatikz[][13B]\unskip\rlap{,}\footnote{\citet{belouadi2024automatikz} refer
to this model as \clima[13B].} the strongest original \automatikz
baseline~\citep{belouadi2024automatikz}, and the general-purpose chatbot
\idefics[8B]~\citep{laurencon2024idefics3}. Additional models and details are available in
\cref{sec:ablations,sec:training-inference-details}.

\paragraph{Results}
We present the system-level metric scores in \cref{tab:adapter-vs-e2e}
(Original Text). On average, \projectname, trained with cosine distance,
achieves the best performance with an AVG score of 85.599, closely followed by
the \mse variant at 85.004. The next best model, \automatikz[v2][\llm], scores
82.375, which is 3 percentage points (pp) lower. The remaining models exhibit a
substantial performance gap, with \idefics and \automatikz[][13B] falling
behind by approximately 20pp and \automatikz[v2][\vlm] showing the weakest
performance across all metrics, resulting in an AVG score of 0. The
surprisingly poor results of \automatikz[v2][\vlm] are likely due to
catastrophic forgetting~\citep{Kirkpatrick2017forgetting}, as the removal of
the vision encoder from \projectname's \vlm necessitates reacquisition of
conditioning based solely on text.

As for individual metrics, our adapter-based models perform particularly well
in perceptual image similarity, with \projectname[Cos] outperforming the best
baseline, \automatikz[v2][\llm], by 3pp on \dsim*. Although
\automatikz[v2][\llm] outperforms \projectname by 1.5pp on \kid, this indicates
in this context that such token-conditioned models (compared to those using
patch embeddings) capture the general appearance of scientific figures well but
fall short in inferring visual specifics from captions. They do, however, have
an edge in reproducing text from captions, which we identify as the primary
reason for up to 5pp higher \clip*, as noted in \cref{sec:clip-matching}.
Regarding code similarity, both \projectname models considerably outperform
others on \cbleu. Interestingly, we observe a mild inverse correlation between
\cbleu and \ted. Models conditioned solely on tokenized captions tend to
generate shorter, often simplified programs~\citep{belouadi2024automatikz},
potentially resulting in a reduced edit distance to the reference. In terms of
efficiency, all models achieve an \mte of 80--85, indicating that only 2 out of
10 inferences require resampling\@. \projectname[Cos] is 3pp more efficient than
\mse, while \automatikz[][13B], likely benefiting from its larger model size,
exceeds it by another 3pp.

In summary, training \automatikz[v2] on top of a \vlm yields worse performance
than training based on vanilla \llama[3.1], indicating that effective
end-to-end training can only leverage the small intersection of graphics
programs and images with captions, as illustrated in \cref{fig:training-data}.
However, even without access to this intersection, \projectname surpasses both
\automatikz[][13B] and \automatikz[v2][\llm] on average by being able to train
on images with captions independently of graphics programs. Moreover, using a
loss function based on cosine distance proves more effective than using \mse.

\subsection{Combining Adapters with Fine-Tuning}\label{sec:adapter-finetuning}
In this section, we investigate whether explicitly incorporating the subset of
\datikz[v3] that includes captions into the training process of \projectname
enhances performance. Our approach involves three incremental stages: (i) We
perform a light fine-tuning of \projectname[Cos] end-to-end on caption-program
pairs for one epoch with a low learning rate of \lr{1}{5}. Extending the
training duration or increasing the learning rate does not yield further
performance gains, likely due to the decoder having already reached its
saturation point; (ii) Alongside feeding captions into the adapter, we provide
them separately to the text decoder in tokenized form (cf.,
\cref{fig:architecture}); (iii) Prior to fine-tuning, we reset the decoder to
its initial weights to overcome saturation, enabling us to fine-tune using the
setup described in \cref{sec:data_model}, which involves 5 epochs and a
learning rate of \lr{5}{5}.

\paragraph{Baselines} In addition to the baselines in
\cref{tab:adapter-vs-e2e}, which remain comparable, we also evaluate larger and
commercial models that serve as stronger baselines (cf.\
\cref{sec:training-inference-details}). Specifically, we assess
\gpt~\citep{openai2023gpt4}, which has demonstrated strong performance in
generating
\tikzname~\citep{bubeck2023sparks,belouadi2024automatikz,zhang2023controllable}
and \qwen~\citep{hui2024qwen25codertechnicalreport} as an open-weights model.

\paragraph{Results} 
In \cref{tab:adapter-ft} (Original Text), all fine-tuning setups of
\projectname show considerable improvement over the base version. Approaches
(i) and (ii) each enhance performance by over 30pp on AVG, while approach (iii)
surpasses them with an improvement of over 70pp, positioning it as the
best-performing model on average, even when compared to our new baselines, with
\gpt being 8pp lower and \qwen approximately 40pp lower. Approach (i)
demonstrates that direct fine-tuning yields positive effects across nearly all
metrics, notably improving \mte by 12pp, \ted by 4pp, and \kid by 3.5pp.
Approach (ii) shows similar trends but, by also incorporating tokenized
captions, further improves \clip* by 5pp, closing the gap to
\automatikz[v2][\llm]. Interestingly, both (i) and (ii) slightly decrease
performance on \cbleu, potentially due to similar reasons discussed in
\cref{sec:adapter-vs-e2e}. However, the same cannot be said for (iii), which
not only achieves the highest score on \cbleu but also ranks as the second-best
on \ted, trailing only 0.5pp behind \gpt and showcasing that it is possible to
perform well on both metrics. Additionally, it increases \dsim* by another 3pp
and \clip* by 8.5pp, competing with the much stronger baselines \qwen and \gpt.
In \kid, it even surpasses them by 3.5pp and 1pp, respectively.

In summary, fine-tuning \projectname, especially when combined with a separate
caption input and weight resetting, greatly improves performance. This
illustrates that the intersection of graphics programs and images with
captions, though small, provides a valuable training signal, and best
performance can be achieved by making full use of both sets. The
best-performing \projectname model even competes with and often surpasses \qwen
and \gpt on several key metrics. Notably, the former model is more than three
times larger, and the latter is often estimated at around 1.8 trillion
parameters~\citep{saxon2024who}, making it 180 times larger.

\subsection{Human Evaluation}\label{sec:human-eval}
To corroborate our findings from automatic evaluation, we conduct a human
annotation campaign focusing on two key properties: caption and image
similarity. We employ \emph{\bws*} (\bws)~\citep{louviere2015bws}, a
comparative annotation method that yields high-quality results even with few
annotators~\citep{kiritchenko2016bws,kiritchenko2017bws}. We sample 100
instances from our test set and present annotators with $n$-tuples of generated
figures, asking them to identify the most and least similar figure to either
the reference caption or reference image. This data is then transformed into
scores from -1 (poor) to 1 (excellent) by subtracting the proportion of times a
figure is selected as the best from the proportion of times it is chosen as the
worst~\citep{orme2009maxdiffa}.
For a manageable workload, we focus on $n=4$ key models: \projectname[Cos], our
best-performing model from \cref{sec:adapter-vs-e2e}; \automatikz[v2][\llm],
its direct end-to-end trained competitor; \gpt, our strongest baseline; and
\projectname[Cos] fine-tuned using approach (iii) from
\cref{sec:adapter-finetuning}, our best model overall, henceforth referred to
as \projectname* for convenience. We engage thirteen annotators and obtain six
fully annotated sets per task (cf.\ \cref{sec:demographics} for more details).
To assess annotator consistency, we calculate the \emph{split-half reliability}
(\shr)~\citep{kiritchenko2017bws}. This method randomly divides all annotations
into two sets, calculates scores independently, and then determines their
correlation using Spearman's $\rho$.

\begin{figure}[tb]
  \centering
  \includegraphics{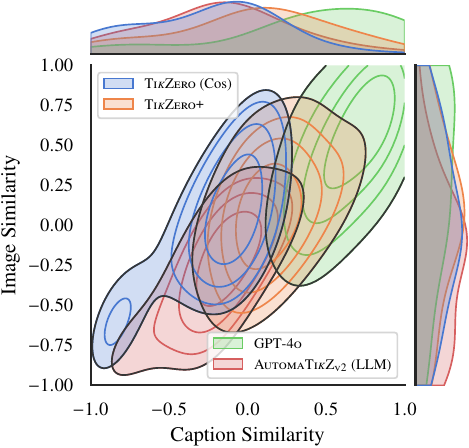}
  \caption{Bivariate distributions of \bws scores (higher is better) using
  kernel density estimation for caption and image similarity. Along the
  diagonal, \projectname[Cos] achieves higher scores than \automatikz[v2][\llm],
  while \projectname* and \gpt demonstrate superior performance compared to
  both.}%
  \label{fig:bwc}
\end{figure}
\paragraph{Results}
\cref{fig:bwc} presents kernel density estimates for the \bws scores, showing
generally consistent rankings with automatic evaluations but revealing notable
differences in the magnitude of gaps. For caption similarity, the ranking
aligns with \clip* evaluations ($\rho=1.0$), with \projectname[Cos],
\automatikz[v2][\llm], \projectname*, and \gpt achieving mean scores $\mu$ of
-0.25, -0.18, 0.03, and 0.4, respectively. Interestingly, humans perceive a
40\% smaller gap between \automatikz[v2][\llm] and \projectname[Cos] than
suggested by \clip* values, where \automatikz[v2][\llm] outperforms
\projectname[Cos] by 50\%. This indicates humans may evaluate caption
similarity differently than \clip* (cf.\ \cref{sec:clip-matching}).
For image similarity, the system order remains consistent with our \dsim*
metric ($\rho=1.0$), with \automatikz[v2][\llm], \projectname[Cos],
\projectname*, and \gpt achieving $\mu$ of -0.26, -0.02, 0.01, and 0.27,
respectively. However, the relative gaps between models differ: the separation
between \automatikz[v2][\llm] and \projectname[Cos], as well as between
\projectname* and \gpt, appear more pronounced than observed with \dsim*. This
discrepancy likely stems from \bws capturing relative preferences rather than
absolute performance differences\@. \gpt is selected 20\% more often as the
best model than \projectname*, and \automatikz[v2][\llm] 15\% more often as the
worst model than \projectname[Cos], creating larger perceived gaps even when
qualitative differences may be subtle.

The \shr values of 0.68 for caption similarity and 0.76 for image similarity
indicate moderate to strong inter-annotator agreement. We also observe a
correlation between these two tasks, with segment-level $\rho=0.62$ and
system-level $\rho=0.8$, suggesting that both evaluation dimensions capture
related aspects of model performance\@. \gpt emerges as the best-performing
model, aligning with its superior performance on the corresponding automatic
metrics, \clip* and \dsim. Among open-source models, \projectname* performs
best, while \automatikz[v2][\llm] ranks lowest overall.

%% file: sections/analysis.tex
\section{Analysis} 
We present a comprehensive analysis, investigating the influence of typographic
attacks on \clip* and examining the effectiveness of our architecture in
low-resource settings, both in terms of training data and trainable parameters.

\subsection{\textbf{\clip*{}} Limitations \& Typographic Attacks}\label{sec:clip-matching}
A known limitation of \clip* with text-rich images is its susceptibility to
typographic attacks, where scores are disproportionately influenced by string
similarity between images and
captions~\citep{ramesh2022dalle2,belouadi2024automatikz}. We suspect that
token-conditioned models like \automatikz[v2][\llm] achieve higher \clip*
values than models such as \projectname[\cosdist \& \mse] primarily because
they tend to visibly copy more substrings from the caption in the output image.
To test this hypothesis, we apply the ROT13 substitution
cipher~\citep{schneier1996cryptography} to all visible strings in the generated
figures and recompute \clip*. This basic cipher replaces each letter with the
13th letter after it in the Latin alphabet. While not cryptographically secure,
the ratio between the original and recomputed \clip* values should indicate the
influence of string matching, i.e., higher ratios suggest less copied text and
vice versa.

\cref{tab:adapter-vs-e2e,tab:adapter-ft} (Redacted Text) present the recomputed
\clip* values and ratios for all evaluated models. The results reveal that most
\projectname models, except for fine-tuning approaches (ii) and (iii), which
also condition on tokenized captions, have considerably higher ratios
(61\%--78\%) compared to strictly token-conditioned models (34\%--51\%),
supporting our hypothesis\@. \automatikz[][13B] is an exception, possibly due to
its initially low score.
Further analysis shows that with redacted text, \automatikz[v2][\llm]'s \clip*
performance drops to the same level as \projectname[\cosdist \& \mse],
suggesting that string matching is the primary factor in its superior
performance rather than producing better visuals---arguably a more difficult
task. Nevertheless, reproducing strings is still somewhat desirable. The human
oracle of our test set achieves a ratio of 50.8\%, close to the 47.5\% of
\projectname*, our best-performing model. In contrast, models like \gpt, with a
lower ratio of 41.9\%, may overfit to caption copying, artificially
inflating the \clip* values.

\subsection{Low-Resource Training}\label{sec:low-resource}
While our adapters train efficiently on large-scale datasets, we investigate
whether such extensive data is necessary for optimal performance. Along the
same vein, we examine the impact of reducing the amount of cross-attention
layers inserted into the vision encoder. We retrain \projectname[\cosdist]
using varying fractions of the training data
($d\in\{1,\frac{1}{2},\frac{1}{4},\frac{1}{8}\}$) and insert cross-attention
layers at different intervals ($i\in\{1,2,4,8\}$).
\begin{table}
  \centering

  \newcolumntype{\avgcol}[1]{>{\collectcell{\gradcol{#1}{92.411}{87.557}{16.25}}}c<{\endcollectcell}}
  \setlength{\extrarowheight}{\belowrulesep}\setlength{\belowrulesep}{0pt}
  \begin{tabular}{c *{4}{\avgcol{2.3}}}
    \toprule
    & \multicolumn{4}{c}{\thead{Training Data}}\\
    \cmidrule(l{\tabcolsep}r{\tabcolsep}){2-5}
    \thead{Intv.} & \multicolumn{1}{c}{100\%} & \multicolumn{1}{c}{50\%} & \multicolumn{1}{c}{25\%} & \multicolumn{1}{c}{12.5\%}\\
    \midrule
    1 & 92.411 & 77.478 & 49.967 & 56.055\\
    2 & 87.557 & 85.249 & 54.942 & 33.817\\
    4 & 82.254 & 47.381 & 32.12  & 37.914\\
    8 & 76.545 & 40.816 & 29.774 & 16.25 \\[-\aboverulesep]
    \bottomrule
  \end{tabular}
  \caption{AVG scores for \projectname[\cosdist] trained on varying fractions
  of data and intervals of cross-attention layers. Higher scores indicate
  better performance. Bold and underlined values denote the best and
  second-best scores for the whole table, respectively. Cell shading
  illustrates score magnitudes.}%
  \label{tab:data-layer-ablation}
\end{table}
\Cref{tab:data-layer-ablation} presents the AVG scores from this parameter
grid, with detailed scores in \cref{sec:experiments-details}. Our findings
reveal that utilizing the full dataset and inserting cross-attention at every
layer yields the highest average performance, highlighting the benefits of
maximizing both variables. Interestingly, the model's performance appears more
robust to a reduction in the number of layers compared to a decrease in
training data. For instance, training on only $\frac{1}{8}$th of the data leads
to a substantial performance drop of 36pp, whereas inserting cross-attention
layers every 8 layers (resulting in only 3 cross-attention layers in total)
causes a more modest decline of 16pp. Minimizing both variables leads to the
most severe drop of over 75pp.
These results validate our training setup while suggesting that incorporating
additional data might further enhance performance. Given that \arxivcap
extracts figures from only 572k papers, whereas some corpora index over 200
million papers~\citep{lo2020s2orc}, there remains a lot of potential for
leveraging larger datasets in future work.

%% file: sections/conclusion.tex
\section{Conclusion} 
In this work, we demonstrate the potential of \projectname and its variants for
generating \tikzname graphics programs from captions. Notably, \projectname
does not require aligned caption-program pairs in its original formulation but
instead aligns representation spaces of unaligned graphics programs and
captioned images. This enables our model to leverage substantially more
training data compared to end-to-end trained models that operate solely on
caption-image pairs (cf.\ \cref{fig:training-data}) while maintaining training
efficiency\@. \projectname outperforms strong end-to-end trained baselines,
including our independently trained \automatikz[v2] models, which use the same
data pool, excluding instances they cannot process, illustrating the strengths
of our approach. When extending the \tikzero approach with additional
end-to-end training, it also compares favorably to much larger baselines and
commercial systems like \gpt. While this enhanced approach, \tikzero*, is no
longer zero-shot by our definition, it remains a \projectname model in the
sense that it operates on both sets of graphics programs and captioned images,
with the added advantage of explicitly utilizing their intersection (cf.\
\cref{fig:training-data}).

These results demonstrate the benefits of designing architectures around
available data and validate the approach of decoupling graphics program
generation from text understanding (with optional later reconciliation through
\tikzero*). Although we demonstrate our method specifically on \tikzname, we
believe its general principles will inspire future work on related graphics
program synthesis tasks.

\paragraph{Future Work}
Beyond scaling up our training data to explore convergence limits (cf.\
\cref{sec:low-resource}), we plan to investigate automatic methods for
improving the quality and alignment of caption-image or caption-program pairs.
This includes rewriting potentially noisy captions with \llm{}s and enhancing
them with the visual understanding capabilities of
\vlm{}s~\citep{nguyen2023improving,awadalla2024blip3,li2024recaptionbillionswebimages}.
We believe our approach to aligning textual and image modalities enables other
promising applications for graphics program synthesis, such as editing images
in latent space via textual instructions to generate modified graphics
programs. Additionally, we intend to explore alternative alignment strategies
beyond model distillation, including contrastive
learning~\citep{Hadsell2006contrastive}, which has successfully aligned
modalities in discriminative
models~\citep{radford2021learningtransferablevisualmodels,girdhar2023imagebind,zhu2024languagebind}.

%% file: sections/limitations.tex
\section*{Limitations} 
Our evaluations include proprietary systems that operate as black boxes; their
training data is unknown, and they offer no guarantees of consistent
performance over time. This (i) makes addressing data leakage and
cross-contamination impossible and (ii) limits the fairness and reproducibility
of our experiments. Nevertheless, even under these unfavorable conditions, our
open models remain competitive. Users should be aware, however, that our models
may behave unpredictably, and outputs might differ from expectations.
Additionally, our models do not include safeguards against potential misuse,
e.g., for generating fake scientific content.

Regarding licensing of our training data, a large portion of the \tikzname
programs in \datikz[v3] are licensed under permissive
terms\footnote{\url{https://creativecommons.org/licenses};
\url{https://opensource.org/license/mit};
\url{https://www.gnu.org/licenses/fdl-1.3.en.html};
\url{https://openai.com/policies/terms-of-use}} that allow redistribution. The
remaining programs are distributed under the \arxiv.org perpetual,
non-exclusive license, which prohibits redistribution, which is why we
\anonymize[will]{} exclude them from the public release of \datikz[v3].
However, since we \anonymize[will]{} release our dataset creation scripts,
we encourage others to reproduce the full version independently.

%% file: sections/acknowledgements.tex
\section*{Acknowledgments}
We extend our sincere gratitude to the following individuals (in no particular
order) for their valuable contributions: Christian Greisinger, Hour Kaing, Ran
Zhang, Tejaswini Medi, Yanran Chen, Sotaro Takeshita, Katharina Prasse, JiWoo
Kim, Christoph Leiter, Haiyue Song, and Aida Kostikova. Their assistance with
our human evaluation campaign, proofreading, insightful discussions, and
constructive feedback has been instrumental to our work.
The first author conducted part of this research during an internship at the
National Institute of Information and Communications Technology (NICT), Japan.
The second to last author is supported by the Federal Ministry of Education and
Research (BMBF) via the research grant \textsc{Metrics4NLG} and the German
Research Foundation (DFG) via the Heisenberg Grant EG 375/5--1.
We acknowledge computing resources provided by the state of Baden-Württemberg
through bwHPC and the German Research Foundation (DFG) through grant INST
35/1597--1 FUGG\@. Finally, we thank the OpenMoji project for the open-source
icons used throughout this work.

%% file: sections/appendix.tex
\section{Additional Baselines and Ablation Studies}\label{sec:ablations} 
Beyond the evaluation in \cref{sec:experiments}, we test additional baselines,
including reasoning models that have proven successful in program synthesis
tasks~\citep{openai2024o1}. We evaluate
\qwen[14B]~\citep{hui2024qwen25codertechnicalreport}, which complements
\qwen[32B] from \cref{sec:adapter-finetuning}, and reasoning models from the
\deepseek[] family (14B and 32B)~\citep{deepseekai2025deepseekr1}. We also
evaluate \projectname[Base], a variant of \projectname[Cos] without the
trainable probe and gating mechanism, to assess their contributions.

As shown in \cref{tab:ablation}, \projectname[Cos] achieves the highest
performance, surpassing \projectname[Base] on both \dsim* and \clip* metrics and
in average performance. These results validate the probe and gate design.
Additionally, \qwen[14B] performs worse than both \projectname[Cos] and, as
expected, its 32B variant in \cref{tab:adapter-ft}. The results are consistent
with our findings in \cref{sec:adapter-vs-e2e} that \projectname[Cos]
outperforms end-to-end trained baselines of comparable size. Notably, the
reasoning models show the lowest overall performance, even compared to
\qwen[14B], indicating that reasoning capabilities alone are insufficient for
graphics program synthesis and more domain-specific post-training may be
needed.
\begin{table*}[b]
 \small
 \centering

 \newcolumntype{\dsimcol}[1]{>{\collectcell{\gradcol{#1}{52.829}{52.373}{44.616}}}c<{\endcollectcell}}
 \newcolumntype{\kidcol}[1]{>{\collectcell{\gradcol{#1}{5.103}{5.225}{15.43}}}c<{\endcollectcell}}
 \newcolumntype{\clipcol}[1]{>{\collectcell{\gradcol{#1}{21.695}{21.201}{9.428}}}c<{\endcollectcell}}
 \newcolumntype{\cbleucol}[1]{>{\collectcell{\gradcol{#1}{1.603}{1.589}{0.229}}}c<{\endcollectcell}}
 \newcolumntype{\tedcol}[1]{>{\collectcell{\gradcol{#1}{60.304}{63.323}{65.51}}}c<{\endcollectcell}}
 \newcolumntype{\mtecol}[1]{>{\collectcell{\gradcol{#1}{93.285}{83.128}{36.11}}}c<{\endcollectcell}}
 \newcolumntype{\avgcol}[1]{>{\collectcell{\gradcol{#1}{64.309}{63.129}{31.102}}}c<{\endcollectcell}}

 \setlength{\extrarowheight}{\belowrulesep}\setlength{\belowrulesep}{0pt}
 \begin{tabular}{l
   \dsimcol{2.3} \kidcol{2.3} \clipcol{2.3} \cbleucol{1.3} \tedcol{2.3} \mtecol{2.3} \avgcol{2.3}}
   \toprule
   \thead{Models} & \nhead{\dsim\up} & \nhead{\kid\down} & \nhead{\clip\up} & \nhead{\cbleu\up} & \nhead{\ted\down} & \nhead{\mte\up} & \nhead{\underline{\avg}\up}\\
   \midrule
   \projectname[Cos] & 52.829 & 5.103 & 10.051 & 1.603 & 65.51 & 82.291 & 64.309\\[-\aboverulesep]
   \midrule
   \projectname[Base] & 52.373 & 5.225 & 9.428 & 1.589 & 65.286 & 83.128 & 63.129\\
   \qwen[14B] & 48.352 & 12.988 & 19.761 & 0.229 & 60.304 & 93.285 & 58.894\\
   \deepseek & 47.573 & 8.887 & 21.201 & 1.388 & 64.928 & 66.225 & 57.252\\
   \deepseek[14B] & 44.616 & 15.43 & 21.695 & 0.842 & 63.323 & 36.11 & 31.102\\[-\aboverulesep]
   \bottomrule
 \end{tabular}
 \caption{System-level $\text{scores}\times100$ for \projectname[Cos] and
 additional baselines. Overall, \projectname achieves the strongest average
 performance across metrics.}\label{tab:ablation}
\end{table*}

\section{Supplementary Comparison with \detikzify}\label{sec:detikzify-details} 
\begin{table*}
  \begin{widetabular}{\textwidth}{l d{2.3} *{2}{d{1.3}} *{3}{d{2.3}} *{2}{d{1.3}} *{2}{d{2.3}}}
    \toprule
    & \multicolumn{5}{c}{\thead{Reference Figures}} & \multicolumn{5}{c}{\thead{Synthetic Sketches}}\\
    \cmidrule(l{\tabcolsep}r{\tabcolsep}){2-6}\cmidrule(l{\tabcolsep}r{\tabcolsep}){7-11}
    \thead{Models} & \nhead{\dsim\up} & \nhead{\kid\down} & \nhead{\cbleu\up} & \nhead{\ted\down} & \nhead{\mte\up}
                   & \nhead{\dsim\up} & \nhead{\kid\down} & \nhead{\cbleu\up} & \nhead{\ted\down} & \nhead{\mte\up}\\
    \midrule
    \detikzify[DS][7b] & 75.46  & 0.842 & 2.953 & 56.851 & 84.019
                       & 67.379 & 0.766 & 1.541 & 59.589 & 84.401\\
    \detikzify[v2]     & \first{2.3}{80.503} & \first{1.3}{0.626} & \first{1.3}{6.105} & \first{2.3}{54.946} & \first{2.3}{93.326}
                       & \first{2.3}{74.584} & \first{1.3}{0.751} & \first{1.3}{3.356} & \first{2.3}{58.32}  & \first{2.3}{93.858}\\
    \bottomrule
  \end{widetabular}
  \caption{System-level $\text{scores}\times100$ for \detikzify[v2] and
  \detikzify[DS][7b] on both reference figures and synthetic sketches
  generated with \ultrasketch from the test split of \datikz[v3]. Best scores
  are in bold, and arrows indicate metric directionality. Note that we compute
  \dsim* using updated models~\citep{sundaram2024when}, whereas
  \citet{belouadi2024automatikz} used the original
  models in their work~\citep{fu2023dreamsim}.}%
  \label{tab:detikzify-comparison}
\end{table*}
\cref{tab:detikzify-comparison} shows in detail how \projectname's inverse
graphics model (hereafter referred to as \detikzify[v2]) compares against
\detikzify[DS][7b], previously the best performing \detikzify model, as
evaluated on the test split of \datikz[v3]\@. \detikzify[v2] clearly
outperforms its predecessor across all evaluated metrics. Below, we briefly
outline key differences in training and inference beyond what we described in
\cref{sec:data_model}. For a comprehensive description of the foundation on
which \detikzify[v2] builds, we refer to \citet{belouadi2024detikzify}.

\paragraph{Training}
Similar to \detikzify, \detikzify[v2] employs a dense layer as the modality
connector between the vision encoder and text decoder. However, for pretraining
this layer, we replace the \metafig dataset~\citep{belouadi2024detikzify} with
the substantially larger \arxivcap dataset, extracting 1 million (figure,
caption, OCR) triplets. During fine-tuning, we randomly substitute inputs with
synthetically generated sketches to support hand-drawn inputs. 
To generate these sketches, we fine-tune the image-editing model
\ultraedit~\citep{zhao2024ultraedit} on a dataset of real, human-created
scientific sketches~\citep{belouadi2024detikzify}. The resulting model,
\ultrasketch, achieves a congruence coefficient (CC)~\citep{sava2006congruence}
of 0.74 with said sketches, compared to 0.72 for the previous model used with
\detikzify. Additionally, we generate synthetic sketches using traditional
image transformations such as random displacement fields. While these sketches
exhibit less diversity, they better preserve text rendering and achieve a
comparable CC of 0.75. Averaging the sketch representations from both methods
increases the CC to 0.82, demonstrating their complementary nature.

\paragraph{Inference}
\detikzify implements a Monte Carlo Tree Search-based inference algorithm
to iteratively refine outputs. As a reward signal $r$, it computes the cosine
similarity $r_\text{cos}=\cos(\operatorname{pool}(\dlvec{x}),
\operatorname{pool}(\dlvec{y}))$ between image patch embeddings
$\dlvec{x},\dlvec{y}$ of input images and compiled outputs via a learned
pooling function. Since \detikzify[v2] fully fine-tunes the vision encoder and
uses its patch embeddings directly, it cannot compute pooled embeddings in the
same way. As an alternative, inspired by popular machine translation
metrics~\citep{belouadi2023uscore,zhao2019moverscore,zhao2020limitations,song2021sentsim},
we experiment with computing the
\emd* (\emd)~\citep{rubner1998emd,kusner2015wmd} with image patch embeddings.
Given the distance matrix $\dlmat{D}$, where $\dlmat[i,j]{D}=\cos(\dlvec[i]{x},
\dlvec[j]{y})$, \emd is defined as follows:
\begin{equation}
  \begin{aligned}
  \operatorname{EMD}(\dlvec{x},\dlvec{y})=&\frac
    {\sum _{i=1}^{\norm{\dlvec{x}}}\sum _{j=1}^{\norm{\dlvec{y}}}\dlmat[i,j]{F}\dlmat[i,j]{D}}
    {\sum _{i=1}^{\norm{\dlvec{x}}}\sum _{j=1}^{\norm{\dlvec{y}}}\dlmat[i,j]{F}},\\
  \text{with}\quad&
    \min\limits_{\dlmat{F}\geq0}{\sum_{i=1}^{\norm{\dlvec{x}}}\sum_{j=1}^{\norm{\dlvec{y}}}\dlmat[i,j]{F}\dlmat[i,j]{D}}\\
  \text{s.t.}\quad&\forall_{i,j}
    \left\lbrace\def\arraystretch{1.2}\begin{array}{@{}l@{}l@{}}
      \sum_{i=1}^{\norm{\dlvec{x}}}\dlmat[i,j]{F} &= \frac{1}{\norm{\dlvec{y}}},\\
       \sum_{j=1}^{\norm{\dlvec{y}}}\dlmat[i,j]{F} &= \frac{1}{\norm{\dlvec{x}}}.
     \end{array}\right.
  \end{aligned}
\end{equation}
When correlating reward scores computed as $r_\text{cos}$ from \detikzify and
$r_\text{EMD}=\operatorname{EMD}(\dlvec[i]{x}, \dlvec[j]{y})$ from
\detikzify[v2] with human judgments from~\citet{belouadi2024detikzify}, we find
that $r_\text{EMD}$ enhances correlation with humans (0.456 segment-level and
0.911 system-level Spearman's $\rho$), compared to $r_\text{cos}$ (0.436 and
0.642, respectively). This demonstrates that \detikzify[v2] not only supports
the inference algorithm but improves upon \detikzify's capabilities.

\section{Supplementary Inference Details}\label{sec:training-inference-details}
To instruct general-purpose models to generate \tikzname code, we employ a
consistent prompt across all models (\gpt, \qwen, and \idefics) originally
engineered by \citet{zhang2025scimage}. For each figure, we replace the
\texttt{<caption>} placeholder with the specific caption:
\begin{prompt}
  Please generate a scientific figure according to the following requirements:
  <caption>. Your output should be in \tikzname code. Do not include any text
  other than the \tikzname code.
\end{prompt}

\section{Supplementary Experimental Results}\label{sec:experiments-details}
\cref{tab:detailed-data-layer-ablation} presents detailed evaluation
metrics scores for the low-resource training experiments discussed in
\cref{sec:low-resource}. The results show a consistent degradation in
performance across all metrics as both the amount of training data and the
number of layers decrease, a trend effectively captured by the AVG scores also
shown in \cref{tab:data-layer-ablation}.
\begin{table*}
  \centering

  \newcolumntype{\dsimcol}[1]{>{\collectcell{\gradcol{#1}{52.771}{52.311}{48.827}}}c<{\endcollectcell}}
  \newcolumntype{\kidcol}[1]{>{\collectcell{\gradcol{#1}{5.103}{5.127}{8.154}}}c<{\endcollectcell}}
  \newcolumntype{\clipcol}[1]{>{\collectcell{\gradcol{#1}{9.955}{9.949}{5.054}}}c<{\endcollectcell}}
  \newcolumntype{\cbleucol}[1]{>{\collectcell{\gradcol{#1}{1.607}{1.509}{1.11}}}c<{\endcollectcell}}
  \newcolumntype{\tedcol}[1]{>{\collectcell{\gradcol{#1}{65.355}{65.399}{66.237}}}c<{\endcollectcell}}
  \newcolumntype{\mtecol}[1]{>{\collectcell{\gradcol{#1}{83.988}{83.679}{77.566}}}c<{\endcollectcell}}
  \newcolumntype{\avgcol}[1]{>{\collectcell{\gradcol{#1}{92.411}{87.557}{16.25}}}c<{\endcollectcell}}

  \setlength{\extrarowheight}{\belowrulesep}\setlength{\belowrulesep}{0pt}
  \begin{tabular}{d{3.3}<{\rlap{\%}} c
    \dsimcol{2.3} \kidcol{2.3} \clipcol{2.3} \cbleucol{1.3} \tedcol{2.3} \mtecol{2.3} \avgcol{2.3}}
    \toprule
    \nhead{Data} &\thead{Intv.} & \nhead{\dsim\up} & \nhead{\kid\down} & \nhead{\clip\up} & \nhead{\cbleu\up} & \nhead{\ted\down} & \nhead{\mte\up} & \nhead{\underline{\avg}\up}\\
    \midrule
     100 & 1 & 52.771 & 5.127 & 9.949 & 1.607 & 65.516 & 82.292 & 92.411\\
     100 & 2 & 52.311 & 5.2   & 9.955 & 1.484 & 65.473 & 82.588 & 87.557\\
     100 & 4 & 51.794 & 5.688 & 8.886 & 1.429 & 65.399 & 83.988 & 82.254\\
     100 & 8 & 51.59  & 5.933 & 9.818 & 1.371 & 65.608 & 83.679 & 76.545\\
      50 & 1 & 52.106 & 5.835 & 8.527 & 1.454 & 65.605 & 83.599 & 77.478\\
      50 & 2 & 52.143 & 5.103 & 9.315 & 1.393 & 65.355 & 82.924 & 85.249\\
      50 & 4 & 50.492 & 6.689 & 8.852 & 1.459 & 65.951 & 78.456 & 47.381\\
      50 & 8 & 50.093 & 6.738 & 7.999 & 1.379 & 65.963 & 78.923 & 40.816\\
      25 & 1 & 51.55  & 6.055 & 9.12  & 1.472 & 66.237 & 77.961 & 49.967\\
      25 & 2 & 51.231 & 6.152 & 8.943 & 1.43  & 65.714 & 77.566 & 54.942\\
      25 & 4 & 49.859 & 7.715 & 7.316 & 1.41  & 66.128 & 79.704 & 32.12 \\
      25 & 8 & 49.179 & 7.764 & 6.495 & 1.434 & 66.009 & 79.9   & 29.774\\
    12.5 & 1 & 50.485 & 6.25  & 7.568 & 1.509 & 65.8   & 80.816 & 56.055\\
    12.5 & 2 & 50.152 & 7.129 & 6.353 & 1.275 & 66.045 & 81.05  & 33.817\\
    12.5 & 4 & 49.667 & 7.031 & 6.474 & 1.221 & 65.892 & 82.634 & 37.914\\
    12.5 & 8 & 48.827 & 8.154 & 5.054 & 1.11  & 65.813 & 80.738 & 16.25 \\[-\aboverulesep]
    \bottomrule
  \end{tabular}
  \caption{System-level $\text{scores}\times100$ \projectname[\cosdist] trained
  on varying fractions of data and intervals of cross-attention layers. Bold
  and underlined values denote the best and second-best scores for the whole
  table, respectively. Cell shading illustrates score magnitudes. Arrows
  indicate metric directionality.}%
  \label{tab:detailed-data-layer-ablation}
\end{table*}

\section{Annotator Demographics}\label{sec:demographics}
Our annotation team consists of thirteen experts with extensive research
experience in Machine Learning, Natural Language Processing, or Computer
Vision. The team includes one male faculty member, four female PhD students,
four male PhD students, and four male researcher scientists from a research
institute. We deliberately selected expert annotators based on findings by
\citet{belouadi2024automatikz}, which demonstrated that crowd workers often
lack the necessary research background to provide reliable annotations for
scientific figures. To mitigate potential biases, each annotator received the
tuples and items within the tuples in randomized order.

\section{Additional Examples}\label{sec:examples}
\begin{figure*}
  \begin{subfigure}[t]{\columnwidth}
    \includegraphics[width=\columnwidth]{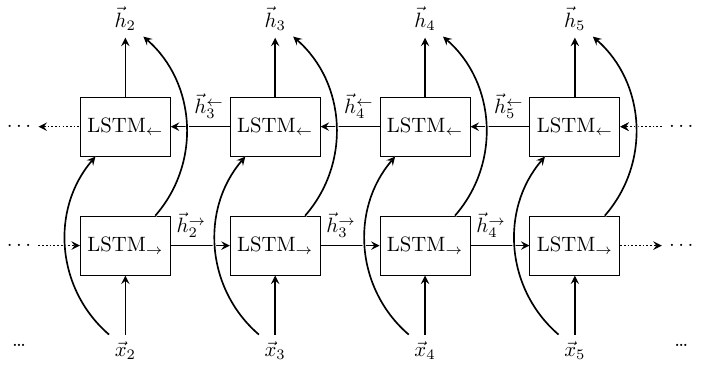}
    \caption{A diagram representing a recurrent neural network consisting
      of several LSTM blocks, processing the input sequence simultaneously
      forwards and backwards (to exploit both directions of temporal
      dependence). Contains some rather tight manoeuvering.}
  \end{subfigure}\hfill%
  \begin{subfigure}[t]{\columnwidth}
    \includegraphics[width=\columnwidth]{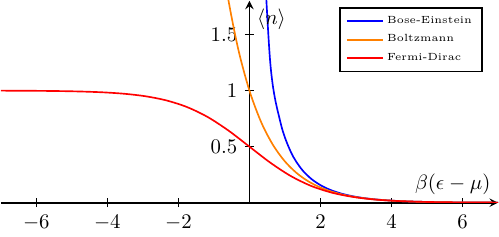}
    \caption{A plot comparing the distribution functions of Bose-Einstein,
    Boltzmann, and Fermi-Dirac statistics as a function of the reduced chemical
    potential $\beta (\epsilon - \mu)$. This visualiation highlights the
    differences between the three types of distribution functions, which are
    used to describe the behavior of particles in different statistical
    systems.}
  \end{subfigure}\medskip

  \begin{subfigure}[t]{\columnwidth}
    \includegraphics[width=\columnwidth]{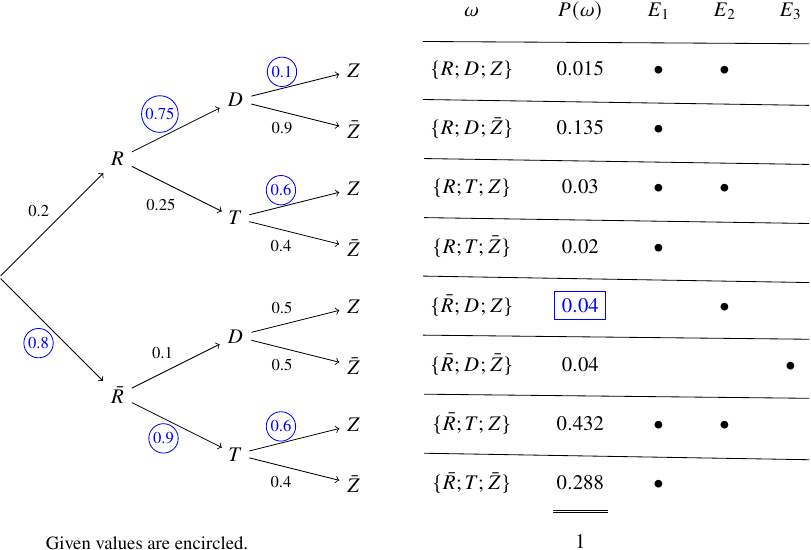}
    \caption{Tree with aligned matrix. A probability tree with an aligned
    matrix listing the possible outcomes, their probabilities and three columns
    for events described in later tasks. It uses the grahdrawing library and
    requires LuaLaTeX.}
  \end{subfigure}\hfill%
  \begin{subfigure}[t]{\columnwidth}
    \includegraphics[width=\columnwidth]{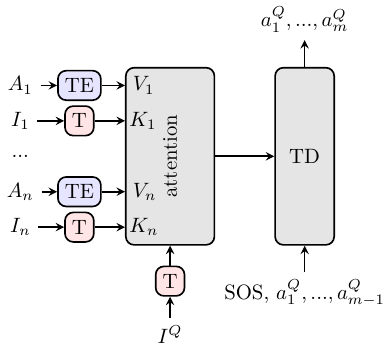}
    \caption{Our approach is a modified version of \textbf{meta-seq2seq}. A
    transformer decoder (TD) is trained to produce a sequence of actions
    $a^Q_1, \ldots, a^Q_{m}$ given a query instruction $I^Q$. The context are
    demonstrations $(I_k, A_k)$ produced by our generative model. We use a
    transformer encoder-decoder (T) to encode instructions and state $S$ and a
    transformer encoder (TE) to encode actions. The transformers that process
    instructions (pink blocks) receive state $S$ as the input of the encoder.}
  \end{subfigure}
  \caption{Representative examples from \datikz[v3] (also present in \datikz and \datikz[v2]), with permissive licenses.}%
  \label{fig:dataset-examples}
\end{figure*}

\Cref{fig:dataset-examples} showcases examples\footnote{sourced from
\url{https://github.com/PetarV-/TikZ}, \url{https://github.com/janosh/tikz},
\url{https://tikz.net}, and \url{https://arxiv.org}} from \datikz[v3] with
permissive licenses. Additionally, \cref{tab:human-eval-exapmles} presents
randomly sampled tuples from our human evaluation with the highest and lowest
rated instances highlighted. The results show that \automatikz[v2][\llm] and
\projectname[Cos] are more frequently selected as the worst models (four and
three times, respectively), while \projectname* and \gpt are more often
chosen as the best models (both three times), which aligns
with our findings in \cref{sec:human-eval}. Finally, \cref{fig:code}
illustrates example programs generated by \projectname* and
\automatikz[v2][\llm], demonstrating how \projectname* utilizes advanced
\tikzname features, whereas \automatikz[v2][\llm] employs only basic, simple
commands.

\begin{table*}
  \newlength\q
  \pgfmathsetlength{\q}{\textwidth/5 -2\tabcolsep}
  \renewcommand\tabularxcolumn[1]{m{#1}}
  \newcommand{\loadfig}[4][]{\raisebox{-.5\ht\strutbox}{%
    \tcbincludegraphics[colback=white,size=small,height=2.25cm,width=\q,#1]{graphics/examples/evaluation/#2/#3/#4_pdf.pdf}}%
    }
  \newcommand{\getcolor}[3]{\noexpandarg\IfEq{#2}{#1}{excel-green}{tcbcolframe}}
  \newcommand{\includerow}[6][caption]{%
    \renewcommand{\ref}[1]{1}%
    \IfStrEq{#1}{caption}%
      {\tiny\input{graphics/examples/evaluation/#1/#2/caption_tex.tex}}%
      {\centering\loadfig[size=minimal]{#1}{#2}{reference}} &
    \loadfig[#3]{#1}{#2}{llama-end-to-end} &
    \loadfig[#4]{#1}{#2}{detikzify-text-adapter-cos} &
    \loadfig[#5]{#1}{#2}{detikzify-text-adapter-cos-reset-captions-ft} &
    \loadfig[#6]{#1}{#2}{gpt-4o}\\
  }
  \setlength{\aboverulesep}{\belowrulesep}
  \begin{tabularx}{\textwidth}{X *{4}{>{\centering\arraybackslash}m{\q}}}
    \toprule
    \nhead{Reference} & \thead{\automatikz[v2]} & \thead{\projectname} & \thead{\projectname*} & \thead{\gpt}\\
    \midrule
        \renewcommand{\ref}[1]{1}    \IfStrEq{caption}{caption}      {\tiny\input{graphics/examples/evaluation/caption/0/caption_tex.tex}}      {\centering\loadfig[size=minimal]{caption}{0}{reference}} &
    \loadfig[colframe=excel-red]{caption}{0}{llama-end-to-end} &
    \loadfig[colframe=excel-red]{caption}{0}{detikzify-text-adapter-cos} &
    \loadfig[]{caption}{0}{detikzify-text-adapter-cos-reset-captions-ft} &
    \loadfig[colframe=excel-green]{caption}{0}{gpt-4o}\\
  
    \midrule
    \includerow[image]{0}{}{colframe=excel-green}{}{colframe=excel-red}
    \midrule
        \renewcommand{\ref}[1]{1}    \IfStrEq{caption}{caption}      {\tiny\input{graphics/examples/evaluation/caption/1/caption_tex.tex}}      {\centering\loadfig[size=minimal]{caption}{1}{reference}} &
    \loadfig[colframe=excel-red]{caption}{1}{llama-end-to-end} &
    \loadfig[]{caption}{1}{detikzify-text-adapter-cos} &
    \loadfig[colframe=excel-green]{caption}{1}{detikzify-text-adapter-cos-reset-captions-ft} &
    \loadfig[]{caption}{1}{gpt-4o}\\
  
    \midrule
    \includerow[image]{1}{colframe=excel-red}{}{colframe=excel-green}{colframe=excel-green}
    \midrule
        \renewcommand{\ref}[1]{1}    \IfStrEq{caption}{caption}      {\tiny\input{graphics/examples/evaluation/caption/2/caption_tex.tex}}      {\centering\loadfig[size=minimal]{caption}{2}{reference}} &
    \loadfig[]{caption}{2}{llama-end-to-end} &
    \loadfig[]{caption}{2}{detikzify-text-adapter-cos} &
    \loadfig[colframe=excel-green]{caption}{2}{detikzify-text-adapter-cos-reset-captions-ft} &
    \loadfig[colframe=excel-red]{caption}{2}{gpt-4o}\\
  
    \midrule
    \includerow[image]{2}{colframe=excel-red}{colframe=excel-green}{}{}
    \midrule
        \renewcommand{\ref}[1]{1}    \IfStrEq{caption}{caption}      {\tiny\input{graphics/examples/evaluation/caption/3/caption_tex.tex}}      {\centering\loadfig[size=minimal]{caption}{3}{reference}} &
    \loadfig[]{caption}{3}{llama-end-to-end} &
    \loadfig[colframe=excel-red]{caption}{3}{detikzify-text-adapter-cos} &
    \loadfig[]{caption}{3}{detikzify-text-adapter-cos-reset-captions-ft} &
    \loadfig[colframe=excel-green]{caption}{3}{gpt-4o}\\
  
    \midrule
    \includerow[image]{3}{colframe=excel-green}{colframe=excel-red}{}{}
    \bottomrule
  \end{tabularx}
  \caption{Alternating rows display randomly selected tuples from the caption
  and image similarity human evaluation task (cf.\ \cref{sec:human-eval}). The
  frames of highest and lowest rated instances are highlighted in
  \textcolor{excel-green}{green} and \textcolor{excel-red}{red},
  respectively.}%
  \label{tab:human-eval-exapmles}
\end{table*}

\begin{figure*}
  \newtcbinputlisting{\examplelisting}[3][]{
    enhanced,%
    size=small,%
    fontupper=\small,%
    left=0pt,%
    right=0pt,%
    top=0pt,%
    bottom=0.3mm-\tcboxedtitleheight,%
    listing only,%
    minted language=latex,%
    minted options={%
      linenos=true,%
      firstnumber=1,%
      numbersep=2mm,%
      breaklines=true,%
      breaksymbol={},%
      breakaftersymbolpre={},%
      breakafter={,},%
      breakindentnchars=4,%
      firstline=1,%
      highlightcolor=nord13,%
      #1,
    },%
    attach boxed title to bottom right={%
      xshift=-0.3mm+0.1pt,
      yshift*=\tcboxedtitleheight+0.3mm-0.1pt%
    },%
    boxed title style={sharp corners=uphill,%
      size=small,%
      no borderline,%
      rightrule=0.1pt,%
      bottomrule=0.1pt%
    },%
    title={#2},%
    listing file={graphics/examples/qualitative/loss/#3_tex.tex},%
  }%
  \centering
  \begin{tcbraster}[raster columns=2,raster equal height=rows]
    \examplelisting{\projectname*}{detikzify-adp/0}
    \examplelisting[numbers=right]{\projectname*}{detikzify-adp/1}
    \examplelisting{\automatikz[v2]}{llama-e2e/0}
    \examplelisting[numbers=right]{\automatikz[v2]}{llama-e2e/2}
  \end{tcbraster}
  \caption{\tikzname programs generated by \projectname* (top) and
  \automatikz[v2][\llm; bottom] corresponding to the figures shown in the first
  row of \cref{fig:example} in the same order.}%
  \label{fig:code}
\end{figure*}

%% file: graphics/examples/evaluation/caption/0/caption_tex.tex
An illustration of the reduction from densest $k$-subgraph to {{\sc u-rcp}}. On the left there is a simple undirected graph $G$ with a single edge. The $2$-reduced directed graph of $G$ is on the right. Each vertex of $G$ is replaced by $2\cdot2 = 4$ copies with a bidirectional edge connecting any two copies of the same vertex, and an outgoing edge from each copy to the single edge-vertex $e$.

%% file: graphics/examples/evaluation/caption/1/caption_tex.tex
Domain of dependence: The closed trapezoidal region $\Omega_t$ used in the proof of {\sc proposition} \ref{prop: finite propagation speed}, shaded with darker yellow; the domain of dependence of point $(0,\ell)$ is $\Omega_t$ along with the top region shaded with lighter yellow.

%% file: graphics/examples/evaluation/caption/2/caption_tex.tex
A sketch of iterating $f(x) = x-1/x$. Points bigger than 1 get sent to point in $[0,1]$ which then moves to a point $\leq -1$ which moves to $[-1,0]$ which is sent to a point $\geq 1$ and so on.

%% file: graphics/examples/evaluation/caption/3/caption_tex.tex
A multiline diagram $D$ with $n=6$ columns and $s=5$ rows, with content $\lambda=(5,4,3,1,0,0)$ and bottom row $\rho^{(1)}(D)=(4,0,1,5,3,0)\in S_\lambda$. It has weight $\operatorname{w}(D)=\operatorname{w}_x(D) \operatorname{w}_t(D)=x_1^3 x_3^2 x_4^4 x_5^2 x_6^2\, t^2$.